\documentclass[10pt,twocolumn,letterpaper]{article}
\usepackage{comment}
\usepackage[pagenumbers]{cvpr} 

\usepackage[accsupp]{axessibility}
\usepackage{graphicx}
\usepackage{amsmath}
\usepackage{amssymb}
\usepackage{booktabs}
\DeclareMathOperator*{\argmin}{arg\,min}

\usepackage{caption,subcaption}
\usepackage[linesnumbered,ruled,vlined]{algorithm2e}
\SetKwInput{KwInput}{Input}
\SetKwInput{KwOutput}{Output}
\usepackage{multirow}
\usepackage[pagebackref,breaklinks,colorlinks]{hyperref}

\usepackage{animate}

\usepackage[capitalize]{cleveref}
\crefname{section}{Sec.}{Secs.}
\Crefname{section}{Section}{Sections}
\Crefname{table}{Table}{Tables}
\crefname{table}{Tab.}{Tabs.}

\def\confName{CVPR}
\def\confYear{2022}

\begin{document}

\title{Vehicle trajectory prediction works, but not everywhere}

\author{
    Mohammadhossein Bahari$^{*,1}$  Saeed Saadatnejad$^{*,1}$ Ahmad Rahimi$^{2}$ Mohammad Shaverdikondori$^{2}$  \\ Amir Hossein Shahidzadeh$^{2}$
    Seyed-Mohsen Moosavi-Dezfooli$^{3}$  Alexandre Alahi$^{1}$ \\  \\
    $^1$EPFL Switzerland \quad $^2${Sharif university of technology} \quad $^3$ Imperial College London \\
    {\tt\small {\{mohammadhossein.bahari, saeed.saadatnejad\}}@epfl.ch}\ \ \\
    }
\maketitle
\let\thefootnote\relax\footnotetext{\leftline{$^*$ Equal contribution as the first authors. }}

\begin{abstract}
Vehicle trajectory prediction is nowadays a fundamental pillar of self-driving cars. Both the industry and research communities have acknowledged the need for such a pillar by providing public benchmarks. While state-of-the-art methods are impressive, \textit{i.e.,} they have no off-road prediction, their generalization to cities outside of the benchmark remains unexplored. In this work, we show that those methods do not generalize to new scenes. We present a method that automatically generates realistic scenes causing state-of-the-art models to go off-road. We frame the problem through the lens of adversarial scene generation. The method is a simple yet effective generative model based on atomic scene generation functions along with physical constraints. Our experiments show that more than $60\%$ of existing scenes from the current benchmarks can be modified in a way to make prediction methods fail (\textit{i.e.,} predicting off-road). We further show that the generated scenes (i) are realistic since they do exist in the real world, and (ii) can be used to make existing models more robust, yielding $30-40\%$ reductions in the off-road rate. The code is available online: \href{https://s-attack.github.io/}{https://s-attack.github.io/}.
\end{abstract}

\section{Introduction}
\begin{figure}[!t]
  \centering
  \includegraphics[width=0.95\columnwidth]{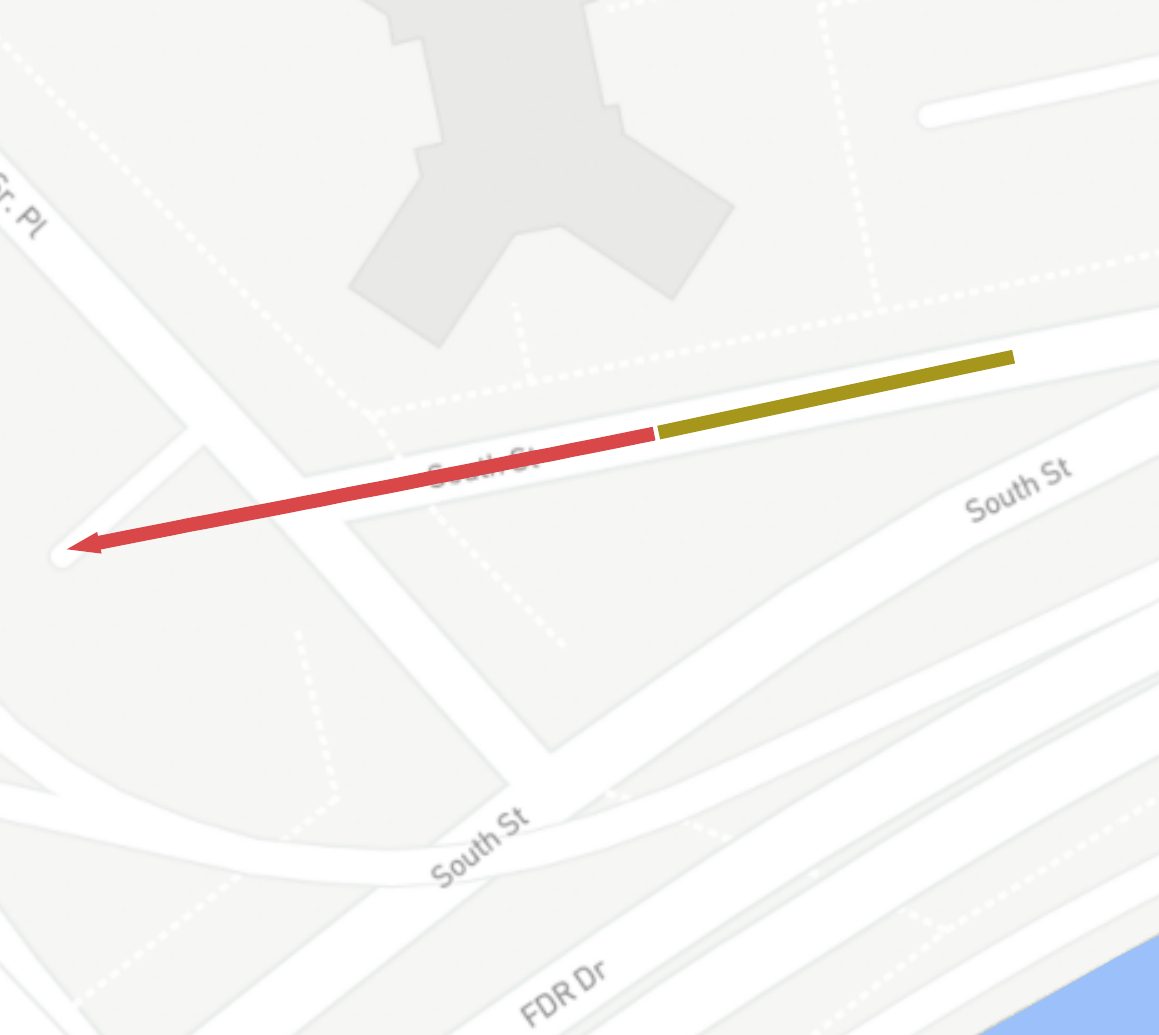}
  \caption{\textbf{A real-world place  (\href{https://www.google.com/maps/@40.7088096,-73.9987943,19.75z}{location}) in New York where the trajectory prediction model (here~\cite{liang2020lanegcn}) fails.} We find this place by retrieving real-world locations which resemble our conditional generated scenes for the prediction model.}
 \label{fig:pull}
\end{figure}

Vehicle trajectory prediction is one of the main building blocks of a self-driving car, which forecasts how the future might unfold based on the road structure (\textit{i.e.,} the scene) and the traffic participants.  
State-of-the-art models are commonly trained and evaluated on datasets collected from a few cities~\cite{ettinger2021large,chang2019argoverse,Caesar2020nuScenesAM}. While their evaluation has shown impressive performance, \textit{i.e.},  almost no off-road prediction, their generalization to other types of possible scenes \textit{e.g.}, other cities, remains unknown.
\Cref{fig:pull} shows a real-world example where a state-of-the-art model reaching zero off-road in the known benchmark~\cite{chang2019argoverse} failed in \textit{South St, New York, USA}. 
Since collecting and annotating data of all real-world scenes is not a viable and affordable solution, we present a method that automatically investigates the robustness of vehicle trajectory prediction to the scene. We tackle the problem through the lens of realistic adversarial scene generation.

Given an observed scene, we want to generate a realistic modification of it such that the prediction models fail in.
Having an off-road prediction is a clear indication of a failure in the the model's scene reasoning and has been used in some previous works~\cite{park2020diverse,casas2020importance,bansal2018chauffeurnet,niedoba2019improving_uber}.
To find a realistic example where the models go off-road, the huge space of possible scenes should be explored.
One solution is data-driven generative models that mimic the distribution of a dataset~\cite{mi2021hdmapgen}. Yet, they do not essentially produce realistic scenes due to the possible artifacts. Moreover, they will represent a portion of real-world scenes as they cannot generate scenes beyond what they have observed in the dataset (cannot extrapolate).
We therefore suggest a simple yet efficient alternative. We show that it is possible to use a limited number of simple functions for transforming the scene into new realistic but challenging ones. Our method 
can explicitly extrapolate to new scenes. 
 
We introduce atomic scene generation functions where given a scene in the dataset, the functions generate multiple new ones.
These functions are chosen such that they can cover a range of realistic scenes.
We then choose the scenes where the prediction model 
produces an off-road trajectory. 
Using three state-of-the-art trajectory prediction models trained on Argoverse public dataset~\cite{chang2019argoverse}, we demonstrate that more than $60\%$ of the existing scenes in the dataset can be modified in such a way that it will make state-of-the-art methods fail (\textit{i.e.,} predict off-road).
We confirm that the generated scenes are realistic by finding real-world locations that partially resemble the generated scenes. We also demonstrate off-road predictions of the models in those locations.
To this end, we extract appropriate features from each scene and use image retrieval techniques to search public maps~\cite{OpenStreetMap}. 
We finally show that these generated scenes can be used to improve the robustness of the models. 

 Our contributions are fourfold:
\begin{itemize}
    \item
    we highlight the need for a more in-depth evaluation of the robustness of vehicle trajectory prediction models;
    \item 
    our work proposes an open-source evaluation framework through the lens of realistic adversarial scene generation by promoting an effective generative model based on atomic scene generation functions;
    \item 
    we demonstrate that our generated scenes are realistic by finding similar real-world locations where the models fail;
    \item 
    we show that we can leverage our generated scenes to make the models more robust.
\end{itemize}


\section{Related work}
\noindent\textbf{Vehicle trajectory prediction.} 
The scene plays an important role in vehicle trajectory prediction as it constrains the future positions of the agents. Therefore, modeling the scene is common in spite of some human trajectory prediction models~\cite{bouhsain2020pedestrian,parsaeifard2021decoupled}. 
In order to reason over the scene in the predictions, some suggested using a semantic segmented map to build circular distributions and outputting the most probable regions~\cite{COSCIA201881}. Another solution is reasoning over raw scene images using convolutional neural networks (CNN)~\cite{lee2017desire}. Many follow-up works represented scenes in the segmented image format and used the learning capability of CNNs over images to account for the scene~\cite{ren2021safety, biktairov2020prank,chai2019multipath,hong2019rules,casas2018intentnet}. Carnet~\cite{sadeghian2018car} used attention mechanism to determine the scene regions that were attended more, leading to an interpretable solution. Some recent work showed that scene can be represented by vector format instead of images~\cite{gao2020vectornet,liang2020lanegcn,song2021learning,bahari2021svg}. 
To further improve the reasoning of the model and generate predictions admissible with respect to the scene, use of symmetric cross-entropy loss~\cite{park2020diverse,Rhinehart_2018_ECCV}, off-road loss~\cite{bansal2018chauffeurnet}, and REINFORCE loss~\cite{casas2020importance} have been proposed.
Despite all these efforts, there has been limited attention to assess the performance of trajectory prediction models on new scenes. Our work proposes a framework for such assessments.

\noindent\textbf{Evaluating self-driving systems.}
Self-driving cars deal with dynamic agents nearby and the static environment around. 
Several works studied the robustness of
self-driving car modules with respect to the status of dynamic agents on the road, \textit{e.g.}, other vehicles. 
Some previous works change the behavior of other agents in the road to act as attackers and evaluate the model's performance with regards to the interaction with other agents~\cite{saadatnejad2021sattack,indaheng2021scenario,abey2019generating,wachi2019failure,2018automatic,2019generating,corso2020survey,kothari2021human}.
Others directly modify the raw sensory inputs to change the status of the agents in an adversarial way~\cite{cao2019adversarial,sun2020towards,tu2020physically,wang2021advsim}. 

In addition to the dynamic agents, driving is highly dependant on the static scene around the vehicle. 
The scene understanding of the models can be assessed by modifying the input scene.
Previous works modify the raw sensory input by changing weather conditions~\cite{tian2018deeptest,zhang2018deeproad,machiraju2020little}, generating adversarial drive-by billboards~\cite{kong2020physgan,zhou2020deepbillboard}, and adding carefully crafted patches/lines to the road~\cite{sato2021dirty,BOLOOR2020101766}.
These works have not changed the shape of the scene, \textit{i.e.}, the structure of the road.
In contrast, we propose a conditional scene generation method to
assess the scene reasoning capability of trajectory prediction models. 
Also our approach is different from data-driven scene generation based on graph~\cite{mi2021hdmapgen} or semantic maps~\cite{saadatnejad2021semdisc}. Data-driven generative models are prone to have artifacts and cannot extrapolate beyond the training data. 
Ours is an adversarial one which can extrapolate to new scenes.

\section{Realistic scene generation}
\label{sec:method}
In this section, we explain in detail our approach for generating realistic scenes. After introducing the notations in \Cref{sec:problem}, we show how we generate each scene in \Cref{sec:scen_gen} and satisfy physical constraints in \Cref{sec:nat_scen}. Finally, we introduce our search method in \Cref{sec:sear_meth}.

\subsection{Problem setup}
\label{sec:problem}
The vehicle trajectory prediction task is usually defined as predicting the future trajectory of a vehicle $z$ given its observation trajectory $h$, status of surrounding vehicles $a$, and scene $S$.
For the sake of brevity, we assume $S$ is in the vector representation format~\cite{chang2019argoverse}$^1$ \footnote{$^1$ We show in \Cref{sec:appendix:raster} that our method is seamlessly applicable when $S$ is in image representation.}. Specifically, $S$ is a matrix of stacked $2d$ coordinates of all lanes' points in $x$-$y$ coordinate space where each row represents a point $s = (s_x, s_y)$.
Formally, the output trajectory $z$ of the predictor $g$ is:
\begin{equation}
    z = g(h, S, a).
\end{equation}
Given a scene $S$, our goal is to create challenging realistic scene $S^*$ 
as we will explain in \Cref{sec:scen_gen}.

\subsection{Conditional scene generation}
\label{sec:scen_gen}

Our controllable scene generation method generates diverse scenes conditioned on existing scenes. Specifically, we opt for a set of atomic functions which represent turn as a typical road topology. 
To this end, we normalize the scene (\textit{i.e.}, translation and rotation with respect to $h$), apply the transformation functions, and finally denormalize to return the generated scene to the original view. Note that every transformation of $S$ is followed by the same transformations on $h$ and $a$. 

\begin{figure*}[!ht]
  \centering
  \begin{subfigure}[b]{0.24\linewidth}
    \centering\includegraphics[width=\linewidth]{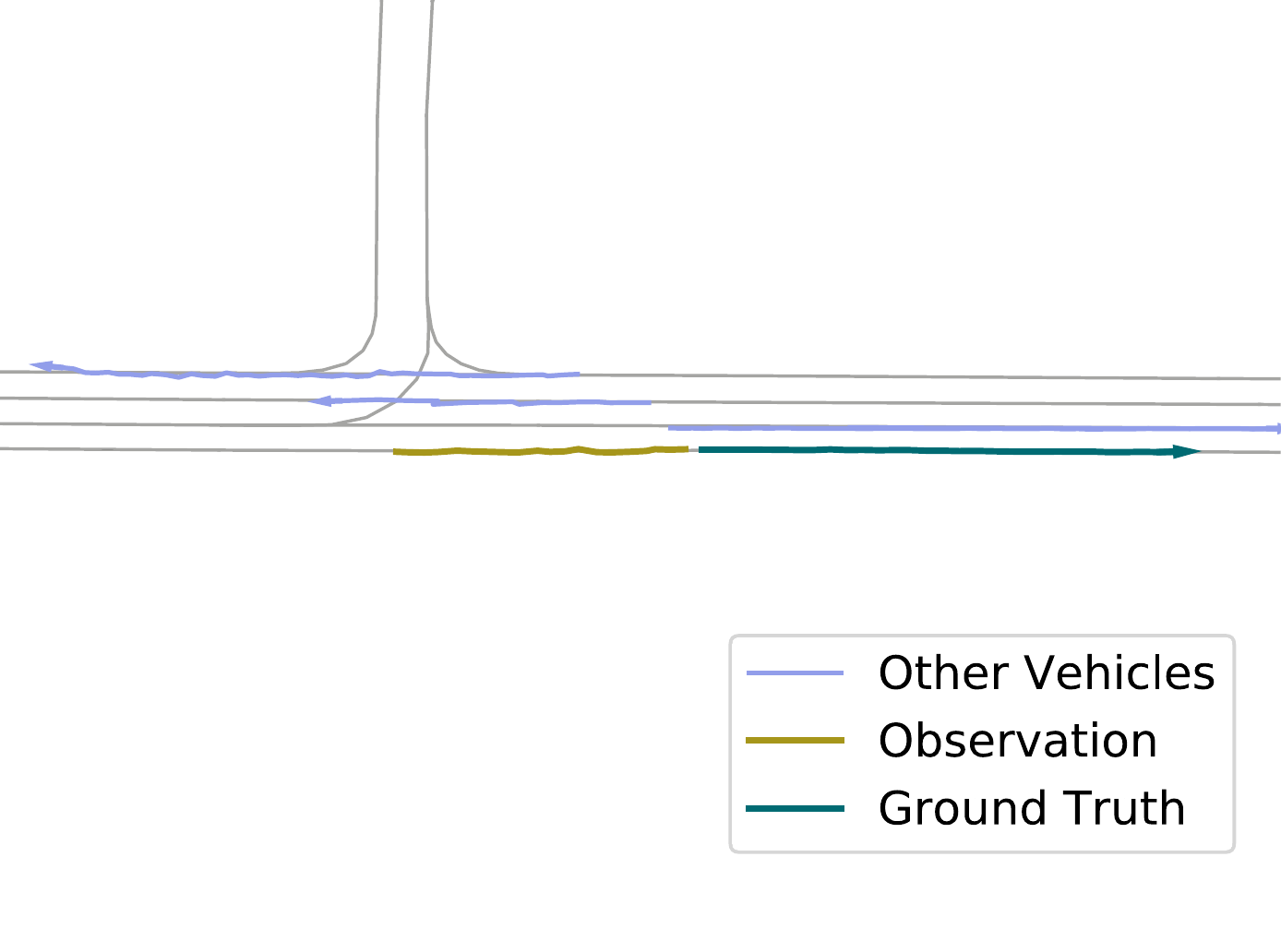}
    \caption{Before transformation}
    \label{fig:attack_methods_normal}
  \end{subfigure}%
  \hfill
    \begin{subfigure}[b]{0.24\linewidth}
    \centering\includegraphics[width=\linewidth]{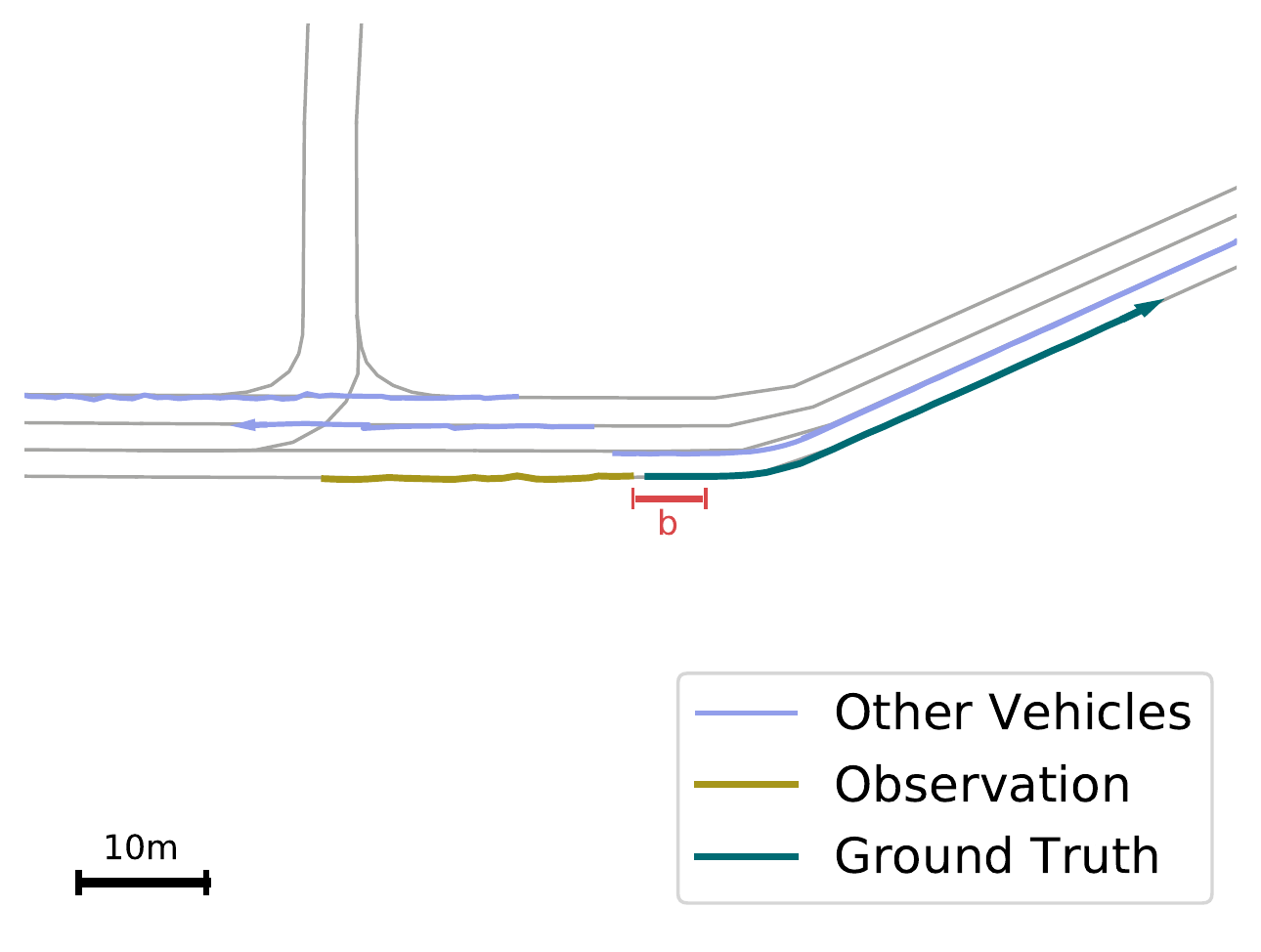}
    \caption{Single-turn}
    \label{fig:attack_methods_single}
  \end{subfigure}%
  \hfill
    \begin{subfigure}[b]{0.24\linewidth}
    \centering\includegraphics[width=\linewidth]{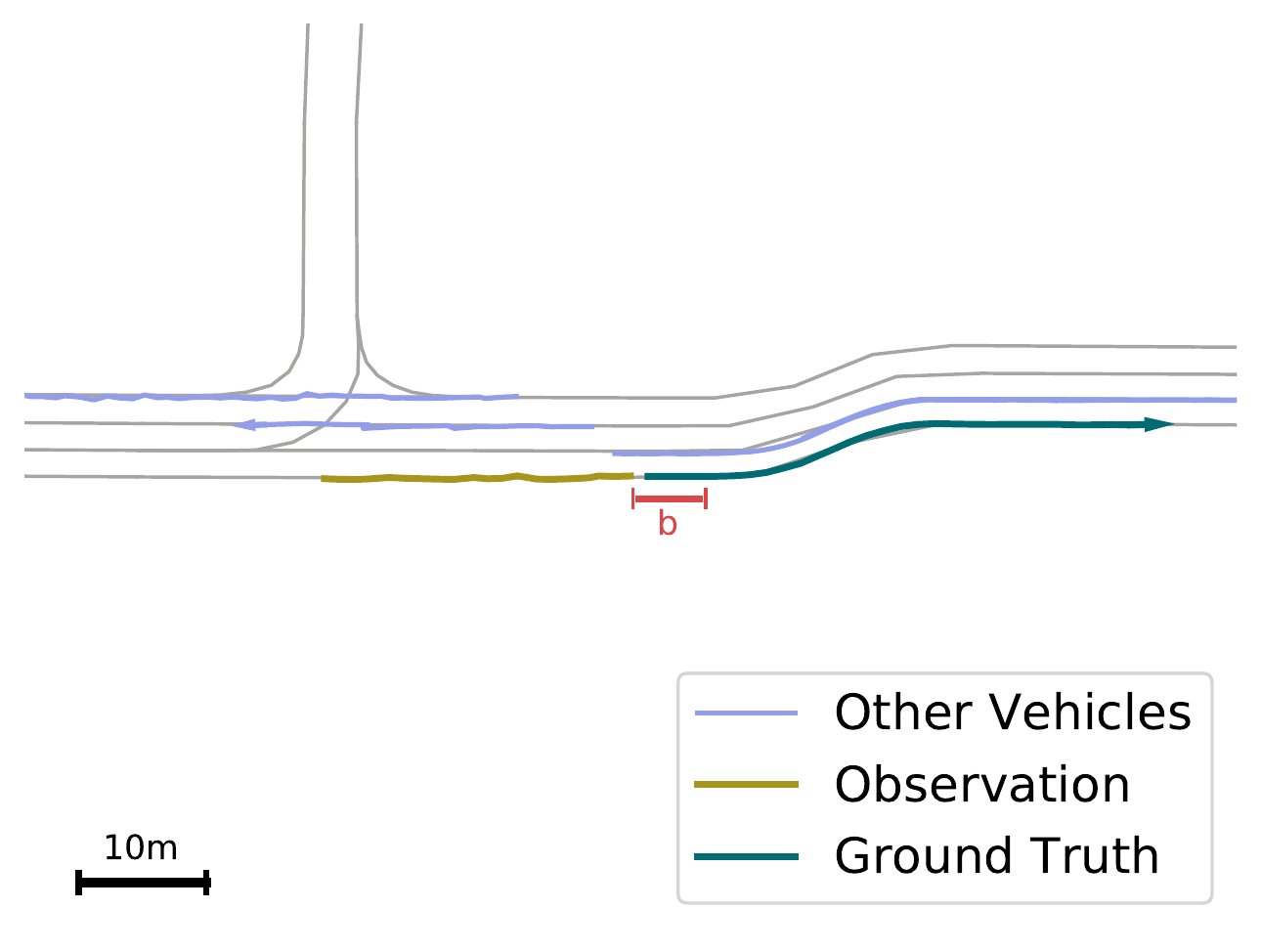}
    \caption{Double-turn}
    \label{fig:attack_methods_double}
  \end{subfigure}
  \hfill
    \begin{subfigure}[b]{0.24\linewidth}
    \centering\includegraphics[width=\linewidth]{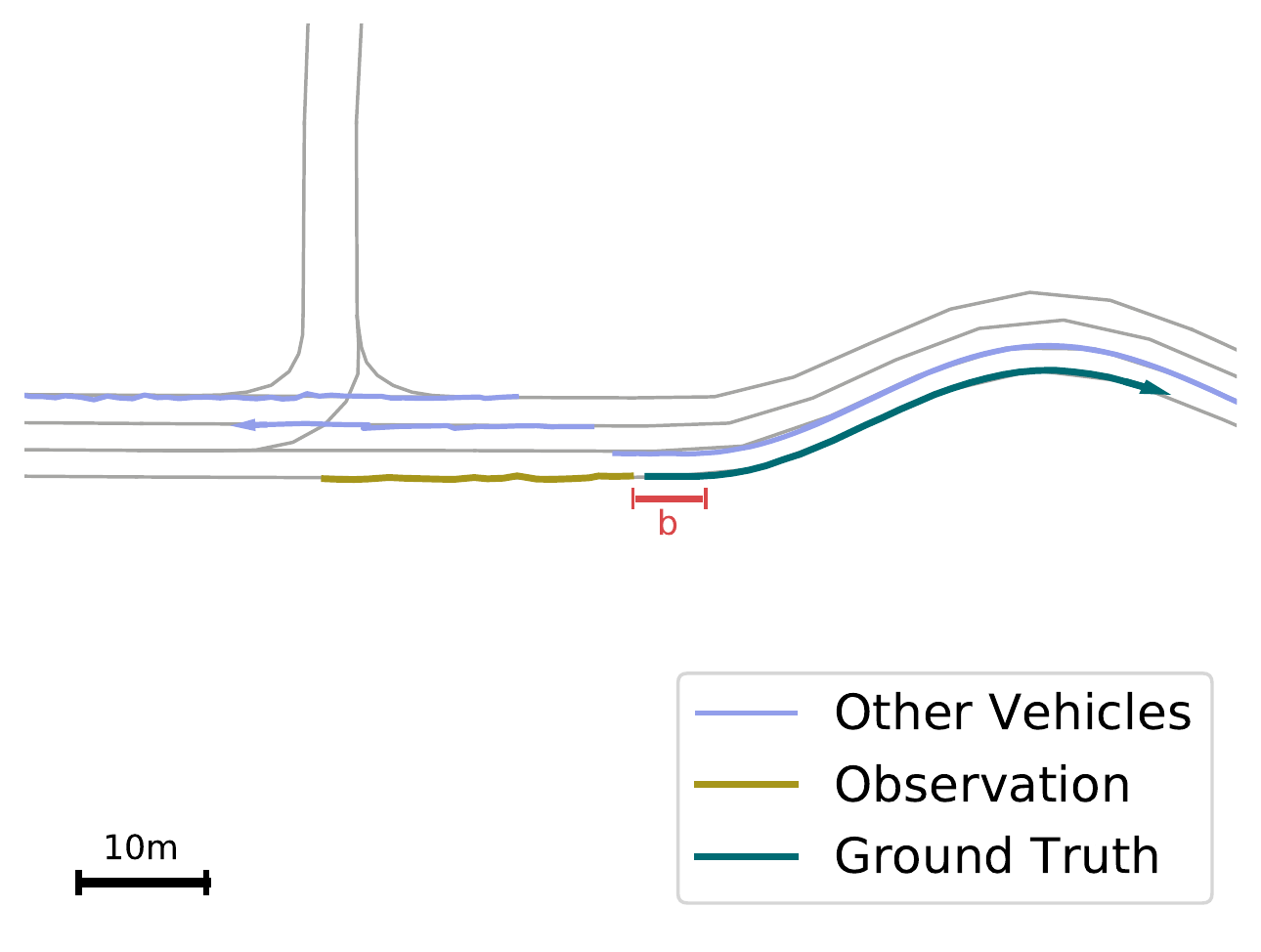}
    \caption{Ripple-road}
    \label{fig:attack_methods_ripple}
  \end{subfigure}
  \caption{\textbf{Visualization of different transformation functions}. The scene before transformation will be followed by three different transformations. Here, $\alpha=(10,0.002,3)$ for the single-turn, $\beta=(10,0.002,3,10)$ for the double-turn and $\gamma=(6,0.017)$ for the ripple-road. $b$ is the border parameter and set to $5$ meters in all figures.}
  \label{fig:attack_methods}
\end{figure*}

We define transformations on each scene point in the following form:
 \begin{equation}
    \tilde{s} = (s_x,s_y+f(s_x-b))
 \label{eq:general_transf}
 \end{equation}
where $\tilde{s}$ is the transformed point, $f$ is a single-variable transformation function, and $b$ is the border parameter that determines the region of applying the transformation.
In other words, we define $f(<0)=0$ so the areas where $s_x<b$ are not modified. This confines the changes to the regions containing the prediction.
One example is shown in \Cref{fig:attack_methods}. 
The new scene is named $\tilde{S}$, a matrix of stacked $\tilde{s}$.
We propose three interpretable analytical functions for the choice of $f$.

\noindent\textbf{Smooth-turn:} this function represents different types of single turns in the road.

\begin{equation}
\begin{split}
     {f_{st, \alpha}}(s_x) &= 
        \begin{cases}
            0,  &s_x < 0\\
            q_{\alpha}(s_x) , &0 \leq s_x \leq \alpha_1 \\
            (s_x-\alpha_1)q'_{\alpha}(\alpha_1)  + q_{\alpha}(\alpha_1) &\alpha_1 < s_x 
        \end{cases},\\
        q_{\alpha}(s_x) &= \alpha_2 s_x^{\alpha_3}, \\
        \alpha &= (\alpha_1, \alpha_2, \alpha_3),
\end{split}
\label{eq:smooth}
\end{equation}
where $\alpha_1$ determines the length of the turn, $\alpha_2, \alpha_3$ control its sharpness, and $q'_{\alpha}$ indicates the derivative of the defined auxiliary function $q_{\alpha}$.
Note that according to the definition, $f_{st,\alpha}$ is continuously differentiable and makes a smooth turn. 
One such turn is depicted in \Cref{fig:attack_methods_single}.

\noindent\textbf{Double-turn:} these functions represent two consecutive turns with opposite directions. Also, there is a variable indicating the distance between them:
\begin{equation}
\begin{split}
    f_{dt,\beta}(s_x) &= f_{st,\beta_1}(s_x) - f_{st,\beta_1}(s_x-\beta_2) , \\ 
    \beta & = (\beta_{11}, \beta_{12}, \beta_{13}, \beta_2 ), \\
    \beta_1 & = (\beta_{11}, \beta_{12}, \beta_{13}),
\end{split}
\label{eq:double}
\end{equation}
where $\beta_1$ is the set of parameters of each turn described in \Cref{eq:smooth} and $\beta_2$ is the distance between two turns. One example is shown in \Cref{fig:attack_methods_double}.

\noindent\textbf{Ripple-road:} one type of scene that can be challenging for the prediction model is ripple road:
\begin{equation}
\begin{split}
    f_{rr, \gamma}(s_x) &= 
        \begin{cases}
            0,&  s_x < 0\\
            \gamma_1(1-cos(2\pi\gamma_2 \ s_x)),&  s_x \geq 0
        \end{cases},    \\
    \gamma &=  (\gamma_1,\gamma_2 ),
\end{split}
\label{eq:ripple}
\end{equation}
where $\gamma_1$ determines the turn curvatures and $\gamma_2$ determines the sharpness of the turns.
One such turn is depicted in \Cref{fig:attack_methods_ripple}.

\subsection{Physical constraints}
\label{sec:nat_scen}
Every scenario consists of a scene and vehicle trajectories in it. The generated scenarios must be feasible, otherwise, they cannot represent possible real-world cases. We consider a scenario as feasible if a human driver can pass it safely. This means that the physical constraints -- \textit{i.e.}, the Newton's law -- should not be violated. The Newton's law indicates a maximum feasible speed for each road based on its curvature~\cite{halliday}:
\begin{equation}
    v_{max} = \sqrt{\mu \textit{g} R},
\end{equation}
where $R$ is the radius of the road, $\mu$ is the friction coefficient and \textit{g} is the gravity.
To consider the most conservative situation, we pick the maximum curvature (minimum radius) existing in the generated road.
Then, we slow down the history trajectory when the speed is higher than the maximum feasible speed, and we name it $\tilde{h}$. 
Note that this conservative speed scaling ensures a feasible acceleration too.
We will show in \Cref{sec:exp} that a model with hard-coded physical constraints successfully predicts the future trajectory for the generated scenes, which indicates that our constraints are enough. 

\subsection{Scene search method}
\label{sec:sear_meth}
In the previous sections, we defined a realistic controllable scene generation method. 
 Now, we introduce a search method to find a challenging scene specific to each trajectory prediction model.

We define $m$ as a function of $z$ and $S$ measuring the percentage of prediction points that are off-road obtained using a binary mask of the drivable area.
We aim to solve the following problem to find a scene in which the prediction model fails in: 
\begin{equation}
\begin{split}
    S^{*} & = \argmin_{\tilde{S}}{l(\tilde{z},\tilde{S})}, \\
    l(\tilde{z},\tilde{S}) &= \left(1 - m(\tilde{z},\tilde{S})\right)^2, 
\label{eq:opt}
\end{split}
\end{equation}
where $\tilde{S}$ is a modification of $S$ according to \Cref{eq:general_transf} using one of the transformation functions \Cref{eq:smooth}, \Cref{eq:double}, or \Cref{eq:ripple}.
Moreover, $\tilde{z} = g(\tilde{h},\tilde{S},\tilde{a})$ is the model's predicted trajectory given the modified scene and the modified history trajectories.
The optimization problem finds the corresponding parameters to obtain $S^*$ that gives the highest number of off-road prediction points.
\Cref{eq:opt} can be optimized using any black-box optimization technique.
We have studied 
Bayesian optimization~\cite{baysian1, baysian2}, Genetic algorithms~\cite{evolution1, evolution2}, Tree-structured Parzen Estimator Approach (TPE)~\cite{parzen} and brute-force.
The overall algorithm is described in \Cref{sec:overal_alg}.


\section{Experiments}
\label{sec:exp}
We conduct experiments to answer the following questions:
1) How is the performance of the prediction models on our generated scenes?
2) Are the generated scenes realistic and possibly similar to the real-world scenes?
3) Are we able to leverage the generated scenes to improve the robustness of the model?

\subsection{Experimental setup}
\subsubsection{Baselines and datasets}
We conduct our experiments on the baselines with different scene reasoning approaches (lane-graph attention~\cite{liang2020lanegcn}, symmetric cross entropy~\cite{park2020diverse}, and counterfactual reasoning~\cite{khandelwal2020whatif}), which
are among the top-performing models and are open-source.

\noindent \textbf{LaneGCN~\cite{liang2020lanegcn}}. It constructs a lane graph from vectorized scene and uses self-attention to learn the predictions.
This method was among the top methods in Argoverse Forecasting Challenge 2020~\cite{argochall_2020}.
It is a multi-modal prediction model which also provides the probability of each mode. Therefore, in our experiments, we consider the mode with the highest probability.

\noindent \textbf{DATF~\cite{park2020diverse}}. It is a flow-based method which uses a symmetric cross-entropy loss to
encourage producing on-road predictions. This multi-modal prediction model does not provide the probability of each mode. We therefore consider the mode which is closest to the ground truth.

\noindent \textbf{WIMP~\cite{khandelwal2020whatif}}. They employ a scene attention module and a dynamic interaction graph to capture geometric and
social relationships. Since they do not provide probabilities for each mode of their multi-modal predictions, we consider the one which is closest to the ground truth.

\noindent \textbf{MPC~\cite{Ziegler2014benz,bahari2021injecting}}.
We report the performance of a rule-based model with satisfied kinematic constraints. We used a well-known rule-based model which follows
center of the lanes~\cite{Ziegler2014benz}. While many approaches can be
used to satisfy kinematic constraints in trajectory prediction, similar to~\cite{bahari2021injecting}, we used Model Predictive Control (MPC) with a bicycle dynamic model.

We leveraged Argoverse dataset~\cite{chang2019argoverse}, the same dataset our
baselines were trained on. Given the $2$ seconds observation trajectory, the goal is to predict the next $3$ seconds as the future motion of the vehicle.
It is a large scale vehicle trajectory dataset. 
The dataset covers parts of Pittsburgh and Miami with total size of 290 kilometers of lanes.

\subsubsection{Metrics}

\noindent \textbf{Hard Off-road Rate} (\textbf{HOR}): in order to measure the percentage of samples with an inadmissible prediction with regards to the scene, we define HOR as the percentage of scenarios that at least one off-road happens in the prediction trajectory points. It is rounded to the nearest integer.
 
\noindent \textbf{Soft Off-road Rate} (\textbf{SOR}): to measure the performance in each scenario more thoroughly, we measure the percentage of off-road prediction points over all prediction points and the average over all scenarios is reported.  The reported values are rounded to the nearest integer.
   
\subsubsection{Implementation details}
We set the number of iterations to $60$, the friction coefficient $\mu$ to $0.7$~\cite{friction} and $b$ equal to $5$ for all experiments.
For the choice of the black-box algorithm, as the search space of parameters is small in our case, we opt for the brute-force algorithm.
We developed our model using a 32GB V100 NVIDIA GPU.

\subsection{Results}
\label{sec:results}
We first provide the quantitative results of applying our method to the baselines in~\Cref{tab:all_methods}. The last column (All) represents the results of the search method described in \Cref{sec:nat_scen}. We also reported the performance of considering only one category of scene generation functions in the optimization problem \Cref{eq:opt} in the other columns of the table. The results indicate a substantial increase in SOR and HOR across all baselines in different categories of the generated scenes. This shows that the generated scenes are difficult for the models to handle.
LaneGCN and WIMP have competitive performances, but WIMP run-time is $50$ times slower than LaneGCN. 
Hence, we use LaneGCN to conduct our remaining experiments.

\begin{figure*}[h]
  \centering
    \begin{subfigure}[b]{0.327\linewidth}
    \centering\includegraphics[width=\linewidth]{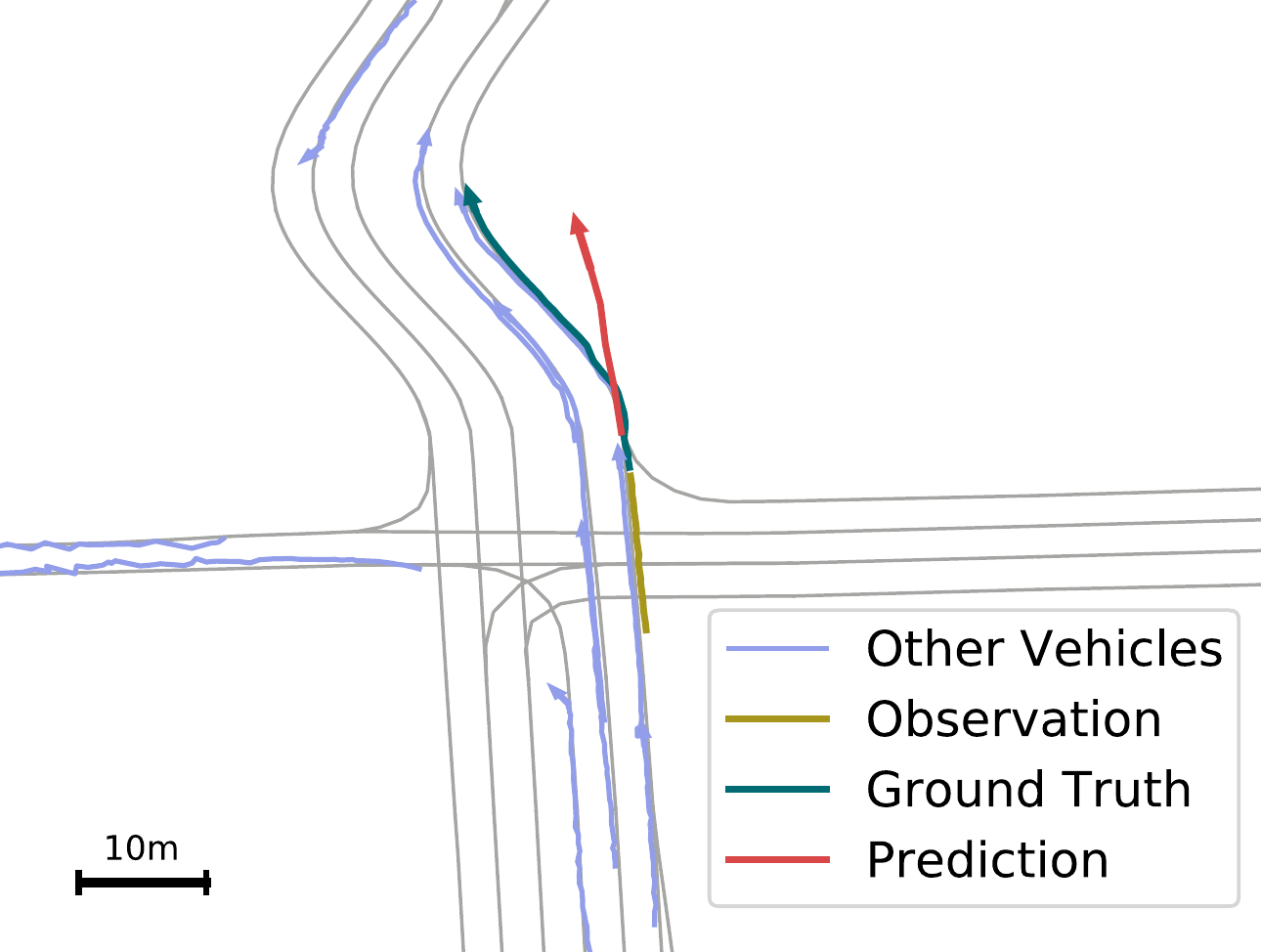}
    \caption{~DATF}
    \label{fig:all_methods_datf}
  \end{subfigure}
  \hfill
  \begin{subfigure}[b]{0.327\linewidth}
    \centering\includegraphics[width=\linewidth]{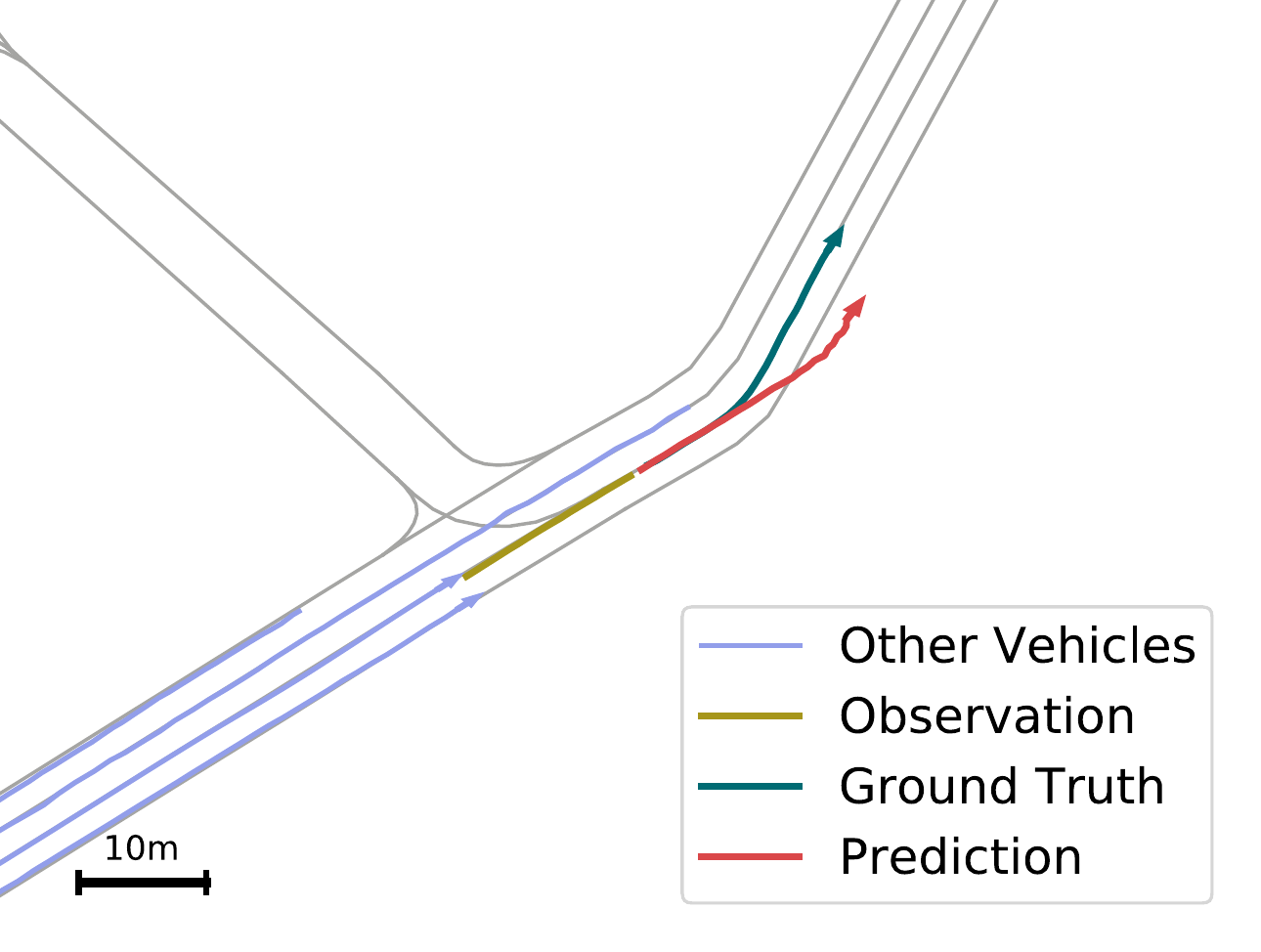}
    \caption{~WIMP}
    \label{fig:all_methods_WIMP}
  \end{subfigure}
  \hfill
  \begin{subfigure}[b]{0.327\linewidth}
    \centering\includegraphics[width=\linewidth]{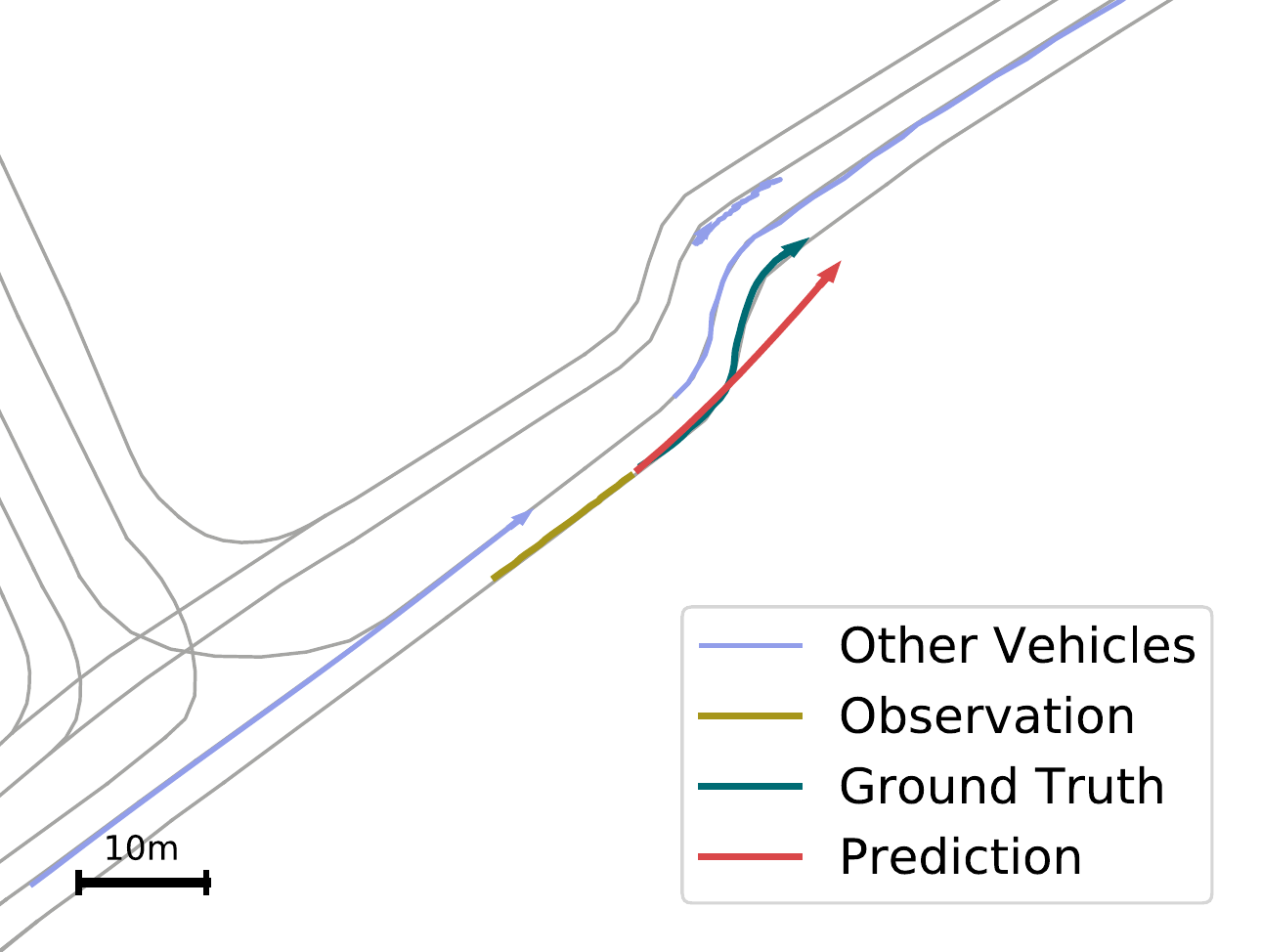}
    \caption{~LaneGCN}
    \label{fig:all_methods_uber}
  \end{subfigure}
  
  \caption{
  \textbf{The predictions of different models in some generated scenes.} All models are challenged by the generated scenes and failed in predicting in the drivable area.
}
 \label{fig:all_methods}
\end{figure*}

\Cref{fig:all_methods} visualizes the performance of the baselines in our generated scenes. We observe that all models are challenged with the generated scenes.
More cases are provided in \Cref{sec:appendix:qual}.

\begin{table*}
\begin{center}
\begin{tabular}{|l|c|c|c|c|c|}
\hline
\multirow{3}{4em}{Model} & Original & \multicolumn{4}{c|}{Generated (\textbf{Ours})} 
\\ 
 &&
 \multicolumn{1}{c|}{Smooth-turn} & \multicolumn{1}{c|}{Double-turn} & \multicolumn{1}{c|}{Ripple-road} & 
 \multicolumn{1}{c|}{All} \\
& SOR / HOR   & SOR / HOR  & SOR / HOR   & SOR / HOR   & SOR / HOR  \\
\hline\hline
DATF~\cite{park2020diverse} & 1 / 2 &  37 / 77 & 36 / 76 & 42 / 80 & 43 / 82\\
WIMP~\cite{khandelwal2020whatif} & 0 / 1 &  13 / 46 & 14 / 50 & 20 / 58 & 22 / 63 \\
LaneGCN~\cite{liang2020lanegcn} & 0 / 1 & 8 / 40 & 19 / 60 & 21 / 62 & 23 / 66 \\
MPC~\cite{Ziegler2014benz} & 0 / 0 & 0 / 0 & 0 / 0 & 0 / 0 & 0 / 0 \\
\hline
\end{tabular}
\caption{\textbf{Comparing the performance of different baselines in the original dataset scenes and our generated scenes.} SOR and HOR are reported in percent and the lower represent a better reasoning on the scenes by the model. 
MPC as a rule-based model always has on-road predictions both in original and our generated scenes.
}
\label{tab:all_methods}
\end{center}
\end{table*}

In \Cref{tab:all_methods}, we observe that SOR is less than or equal to $1\%$ for all methods in the original scenes.
Our exploration shows that more than $90\%$ of these off-road cases are due to the annotation noise in the drivable area maps of the dataset and the models are almost error-free with respect to the scene. Some figures are provided in \Cref{sec:appendix:qual}.
While this might lead to the conclusion that the models are flawless, results on the generated scenes question this conclusion.
We confirm our claim in the next section by retrieving the real-world scenes where the model fails.

Feasibility of a scenario is an important feature for generated scenes. As mentioned in \Cref{sec:nat_scen}, we added physical constraints to guarantee the physical feasibility of the scenes. \Cref{tab:all_methods} indicates that MPC as a rule-based model predicts almost without any off-road in the generated scenarios. It approves that the scenes are feasible with the given history trajectory.
In order to study the importance of added constraints, we relax the constraints for the generated scenes.
We report the performance of the baseline and MPC on the cases where the maximum speed in their $h$ is higher than $v_{max}$.
In \Cref{tab:mpc}, we observe that without those feasibility-assurance constraints, there are more cases where MPC is unable to follow the road and has $3 \times$ more off-road. We conclude that those constraints are necessary to make the scene feasible. 
We keep the constraints in all of our experiments to generate feasible scenarios.

\begin{table}
\centering
\begin{tabular}{|l|c|c|}
\hline
\multirow{2}{4em}{Model} & 
\multicolumn{1}{c|}{w/ phys} & \multicolumn{1}{c|}{w/o phys } \\
& SOR / HOR & SOR / HOR  \\
\hline\hline
LaneGCN & 
33 / 85 & 47 / 92 \\
MPC &
0 / 1 & 0 / 3 \\

\hline
\end{tabular}
\begin{center}
\caption{\textbf{Impact of the physical constraints.} We report the performance with and without the physical constraints explained in \Cref{sec:nat_scen}. The numbers are reported on samples of data with speed higher than $v_{max}$ in their $h$.
}
\label{tab:mpc}
\end{center}
\end{table}

\subsection{Real-world retrieval}
\begin{figure*}[h]
  \centering
  \begin{subfigure}[b]{0.327\linewidth}
    \centering\includegraphics[width=\linewidth]{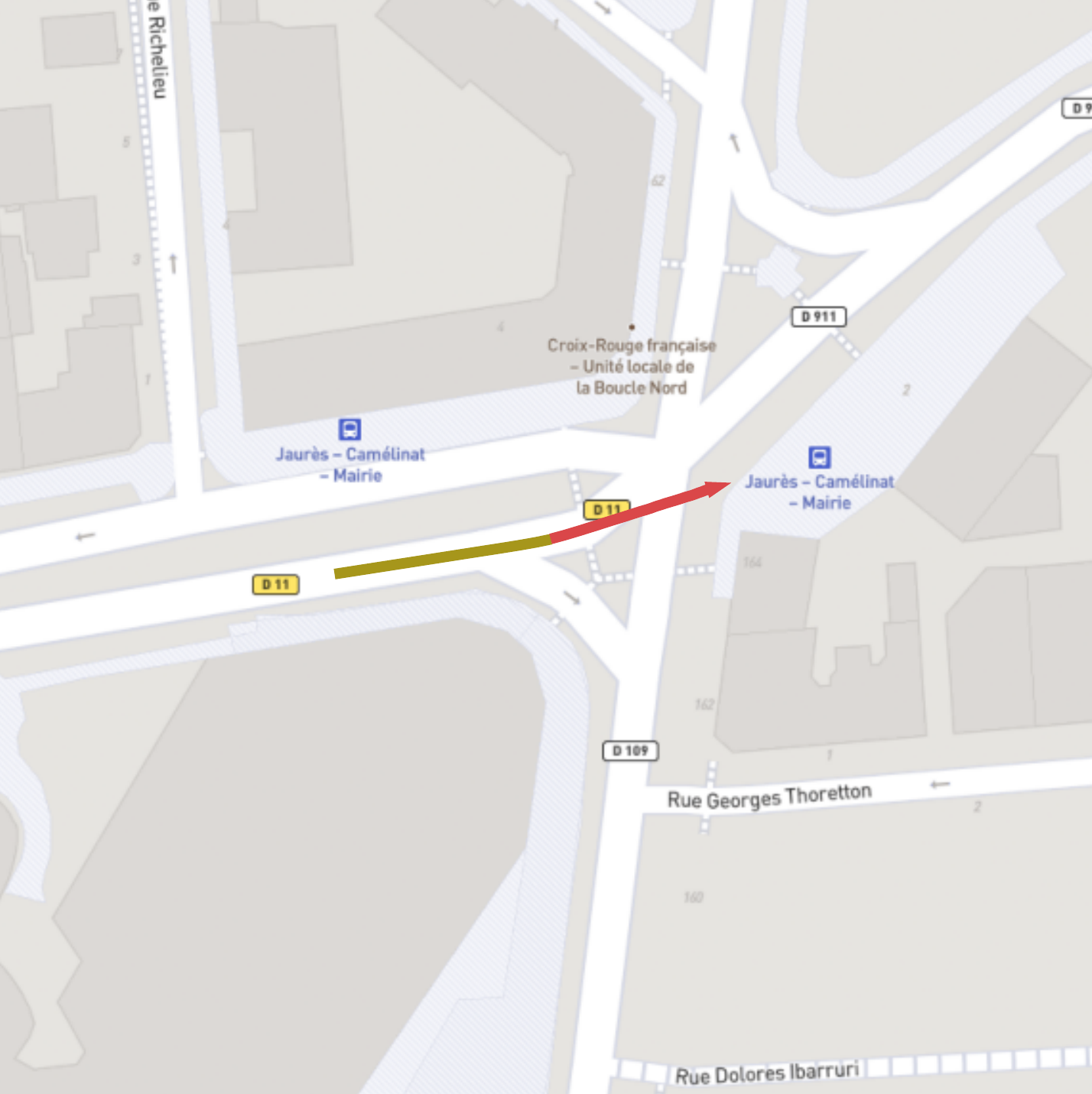}
    \caption{Paris \href{https://www.google.com/maps/@48.9267618,2.2944925,19z}{location}}
  \end{subfigure}%
  \hfill
  \begin{subfigure}[b]{0.327\linewidth}
    \centering\includegraphics[width=\linewidth]{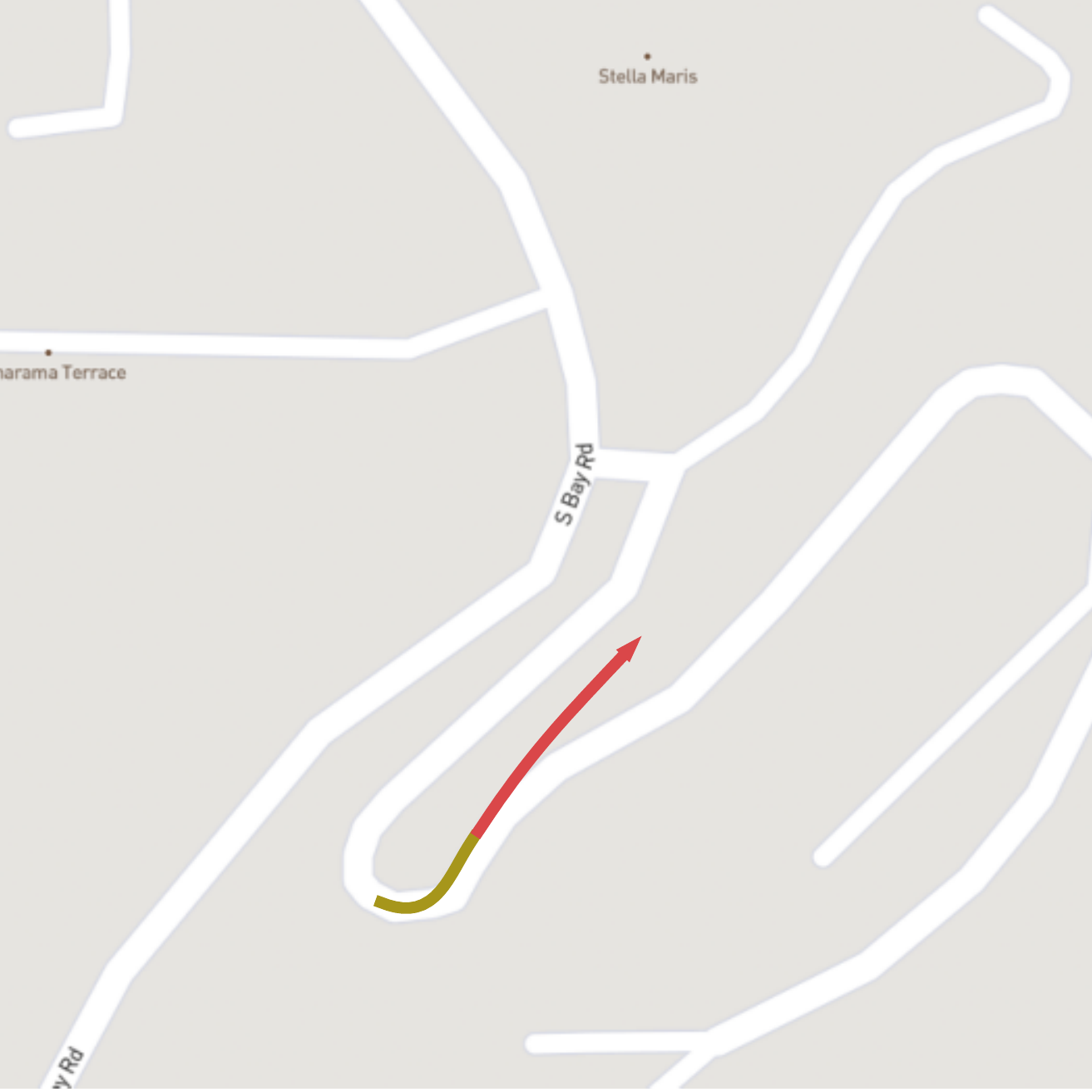}
    \caption{Hong Kong \href{https://www.google.com/maps/@22.2332883,114.1992906,20.25z}{location}}
  \end{subfigure}
  \hfill
    \begin{subfigure}[b]{0.327\linewidth}
    \centering\includegraphics[width=\linewidth]{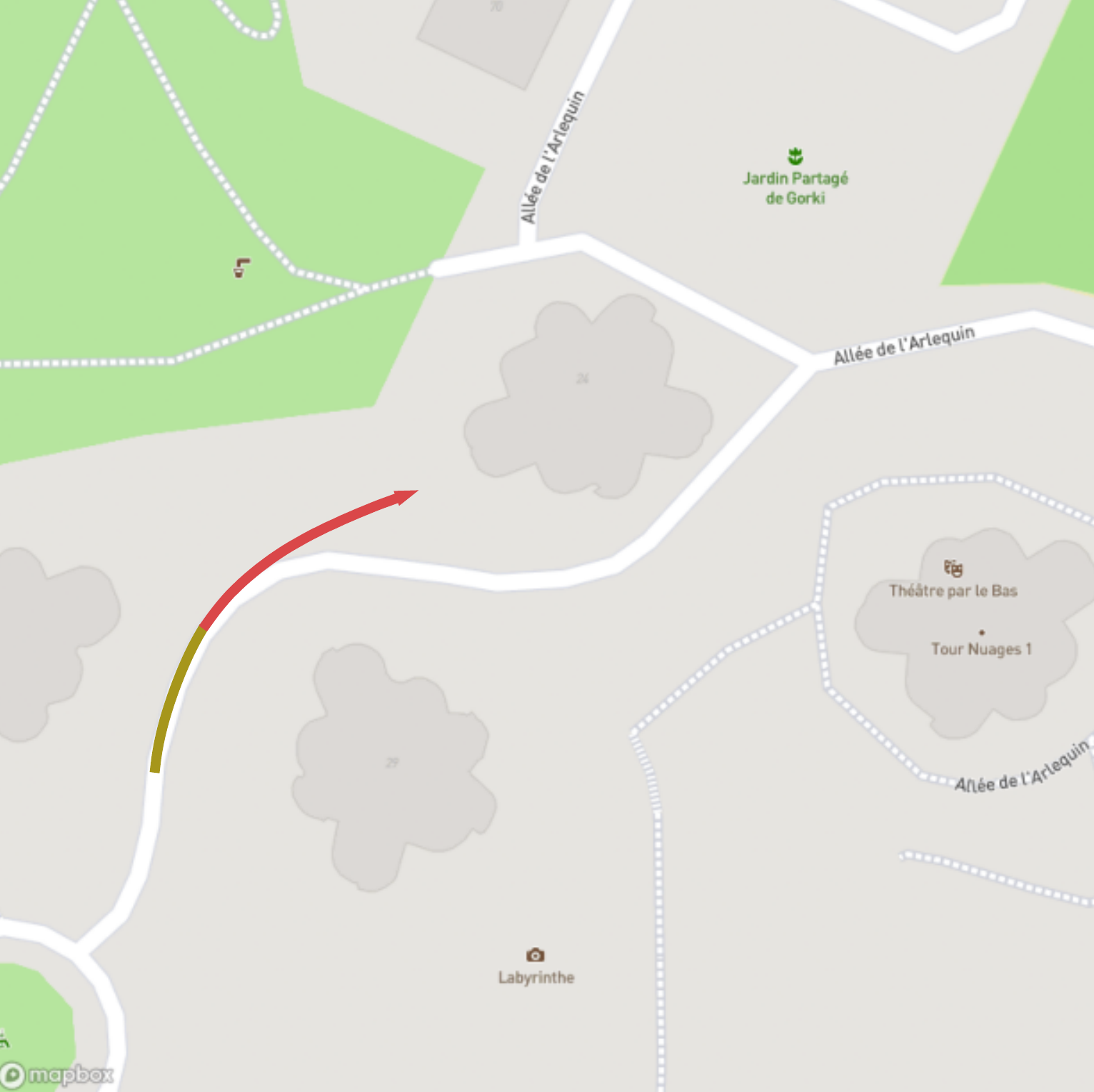}
    \caption{New Mexico \href{https://www.google.com/maps/@48.8905404,2.2268305,19.25z}{location}}
  \end{subfigure}
  \caption{
  \textbf{Retrieving some real-world locations similar to the generated scenes using our real-world retrieval algorithm.} We observe that the model fails in Paris (a), Hong Kong (b) and New Mexico (c).}
 \label{fig:real-world}
\end{figure*}

So far, we have shown that the generated scenes along with the constraints are feasible/realistic scenes.
Next, we want to study the plausibility/existence of the generated
scenes.
Inspired by image retrieval methods~\cite{nister2006ktree}, we develop a retrieval method to find similar roads in the real-world.
First, we extract data of $4$ arbitrary cities (New York, Paris, Hong Kong, and New Mexico) using OSM~\cite{OpenStreetMap}. Then, $20,000$ random samples of $200 \times 200$ meters are collected from each city. Note that it is the same view size as in Argoverse samples.
Then, a feature extractor is required to obtain a feature vector for each scene. 
We used the scene feature extractor of LaneGCN named MapNet to obtain some $128$ dimensional feature vectors for each sample. 
We then use the well-known image retrieval method K-tree algorithm~\cite{nister2006ktree}. It first uses K-Means algorithm multiple times to cluster the feature vectors 
of all scenes into a predefined number of clusters (in our case $1000$). 
Then, given a generated scene as the query, it sorts real scenes based on the similarity with the query scene
and retrieves $10$ closest scenes to the query. 
Finally, we test the prediction model in these examples. Some examples are provided in \Cref{fig:real-world}. More scenes can be found \Cref{sec:appendix:qual}.

\subsection{Robustness}
Here, we study if we can make the models robust against new generated scenes.
To this end, we fine-tune the trained model using a combination of the original training data and the generated examples by our method for $10$ epochs. 

We report the performance of these models in the generated scenes with different transformation power. Transformation power is determined by $\alpha_2 \times 3000$, $\beta_{12} \times 3000$ and $\gamma_1$ for \Cref{eq:smooth}, \Cref{eq:double}, and \Cref{eq:ripple}, respectively. It represents the amount of curvature in the scene. 
\Cref{tab:robustness} indicates that without losing the performance in the original accuracy metrics, the fine-tuned model is less vulnerable to the generated scenes by predicting $40\%$ less SOR and $30\%$ less HOR in the Full setting.
While the results show improvements in all transformation powers, the gains in extreme cases are higher, \textit{i.e.}, the model can handle them better after fine-tuning.

\begin{table*}[!t]
\begin{center}
\begin{tabular}{|l|ccccc|}
\hline
\multirow{2}{4em}{Model}
& Pow=1 & Pow=3 & Pow=5 & Pow=7 & Pow=9 (Full) \\
& SOR/HOR & SOR/HOR & SOR/HOR & SOR/HOR & SOR/HOR  \\
\hline\hline
LaneGCN & 2 / 8 & 12 / 35 & 19 / 49 & 22 / 58 & 23 / 66 \\
LaneGCN w/ aug & \textbf{1 / 7} & \textbf{6 / 21} & \textbf{10 / 30} & \textbf{13 / 38} & \textbf{14 / 46} \\
\hline
\end{tabular}
\caption{\textbf{Comparing the original model and the fine-tuned model with data augmentation of the generated scenes.} The performance is reported on generated scenes with different transformation power (Pow). Transformation power is determined by $\alpha_2 \times 3,000$, $\beta_{12} \times 3,000$ and $\gamma_1$ for \Cref{eq:smooth}, \Cref{eq:double}, and \Cref{eq:ripple}, respectively which represents the amount of curvature in the scene.
The average / final displacement errors on original scenes are equal to $1.35 / 2.98 m$ for both original and fine-tuned models. 
}
\label{tab:robustness}
\end{center}
\end{table*}

\begin{figure}[!t]
 \centering
 \begin{subfigure}[b]{0.49\columnwidth}
    \centering    \includegraphics[width=\linewidth]{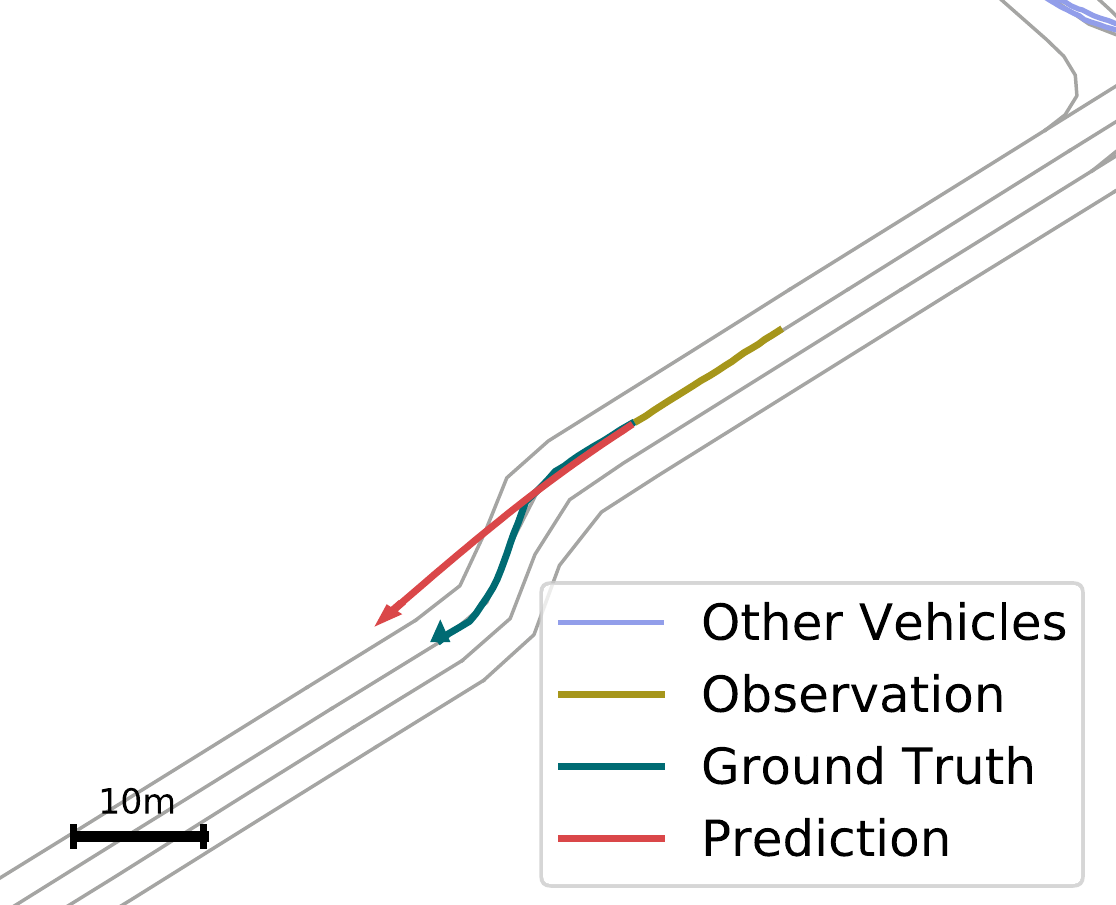}
    \label{fig:before_robust}
    \end{subfigure}
    \hfill
    \begin{subfigure}[b]{0.49\columnwidth}
    \centering    \includegraphics[width=\linewidth]{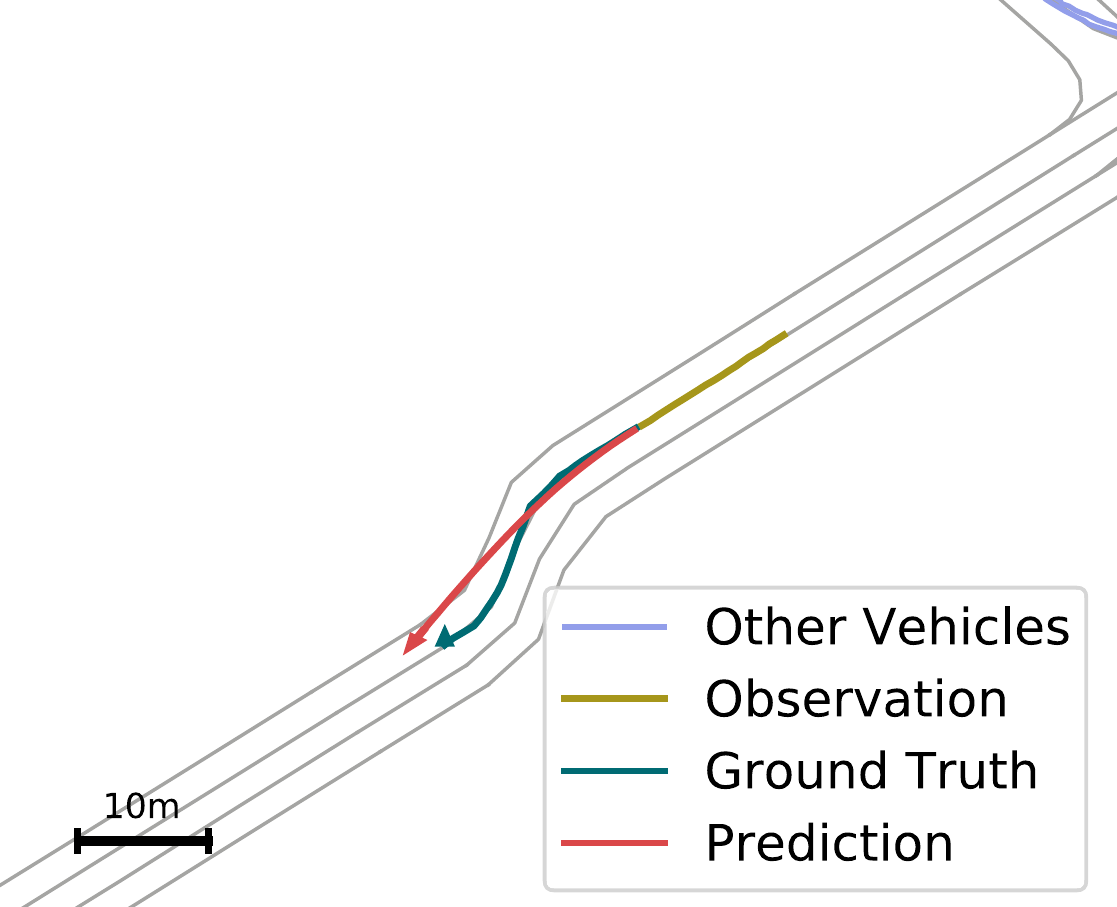}
    \label{fig:after_robust}
    \end{subfigure}
 \caption{\textbf{The output of the original model (the left) vs the robust model (the right) in a generated scene.} While the original model has a trajectory in non-drivable area, the robust model predicts without any off-road.}
 \label{fig:robust}
\end{figure}

In \Cref{fig:robust}, the prediction of the original model is compared with the prediction of the robust model. 
The original model cannot predict without off-road while the fine-tuned model is able to predict reasonable and without any off-road point.

\subsection{Discussions}
In this section, we perform experiments and bring speculations to shed light on the weaknesses of the models. 

\begin{figure}[!t]
    \centering
    \begin{subfigure}[b]{0.49\columnwidth}
    \includegraphics[width=\linewidth]{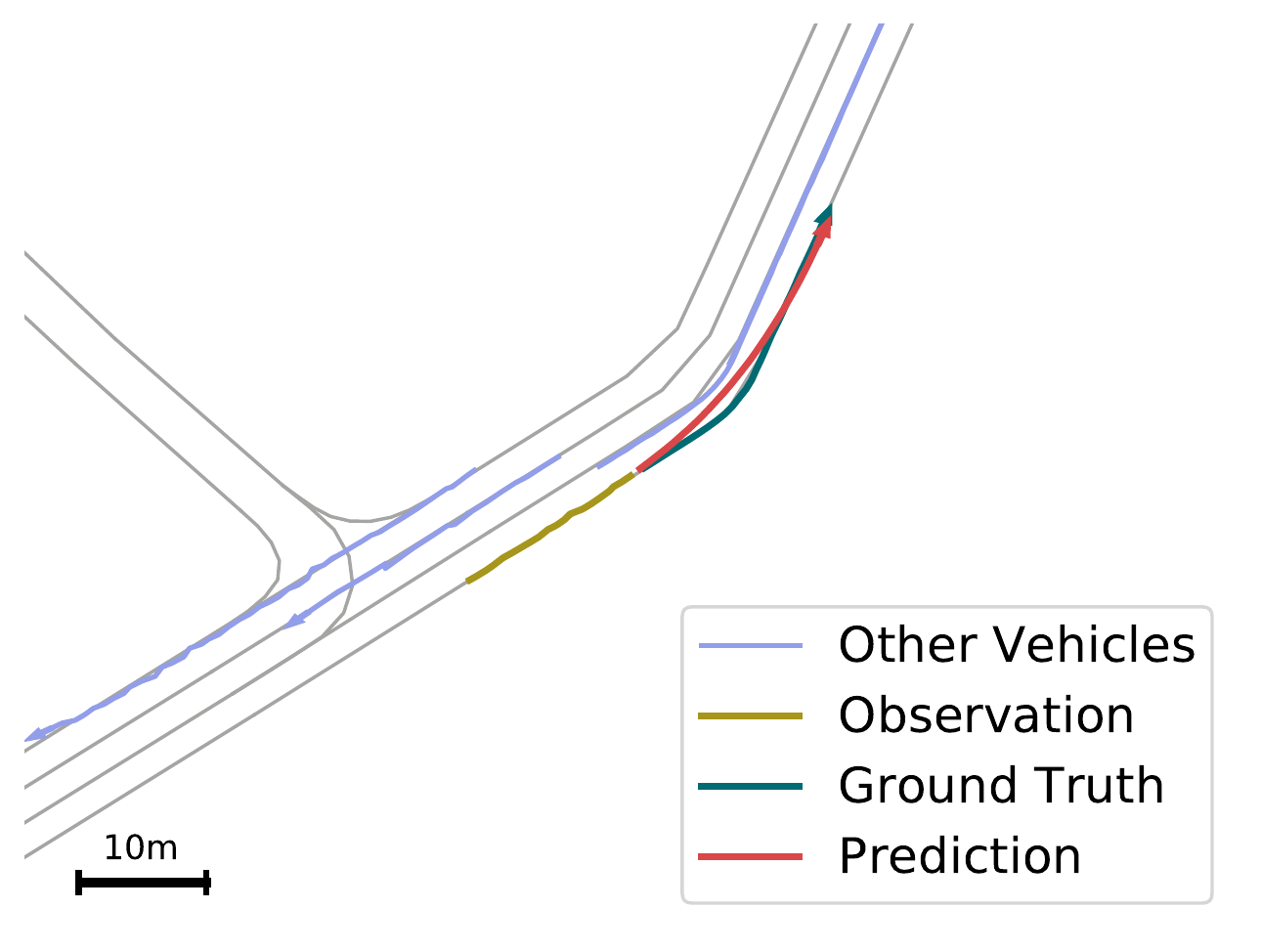}
    \caption{ }
    \label{fig:failure1}
    \end{subfigure}
    \hfill
    \begin{subfigure}[b]{0.49\columnwidth}
    \includegraphics[width=\linewidth]{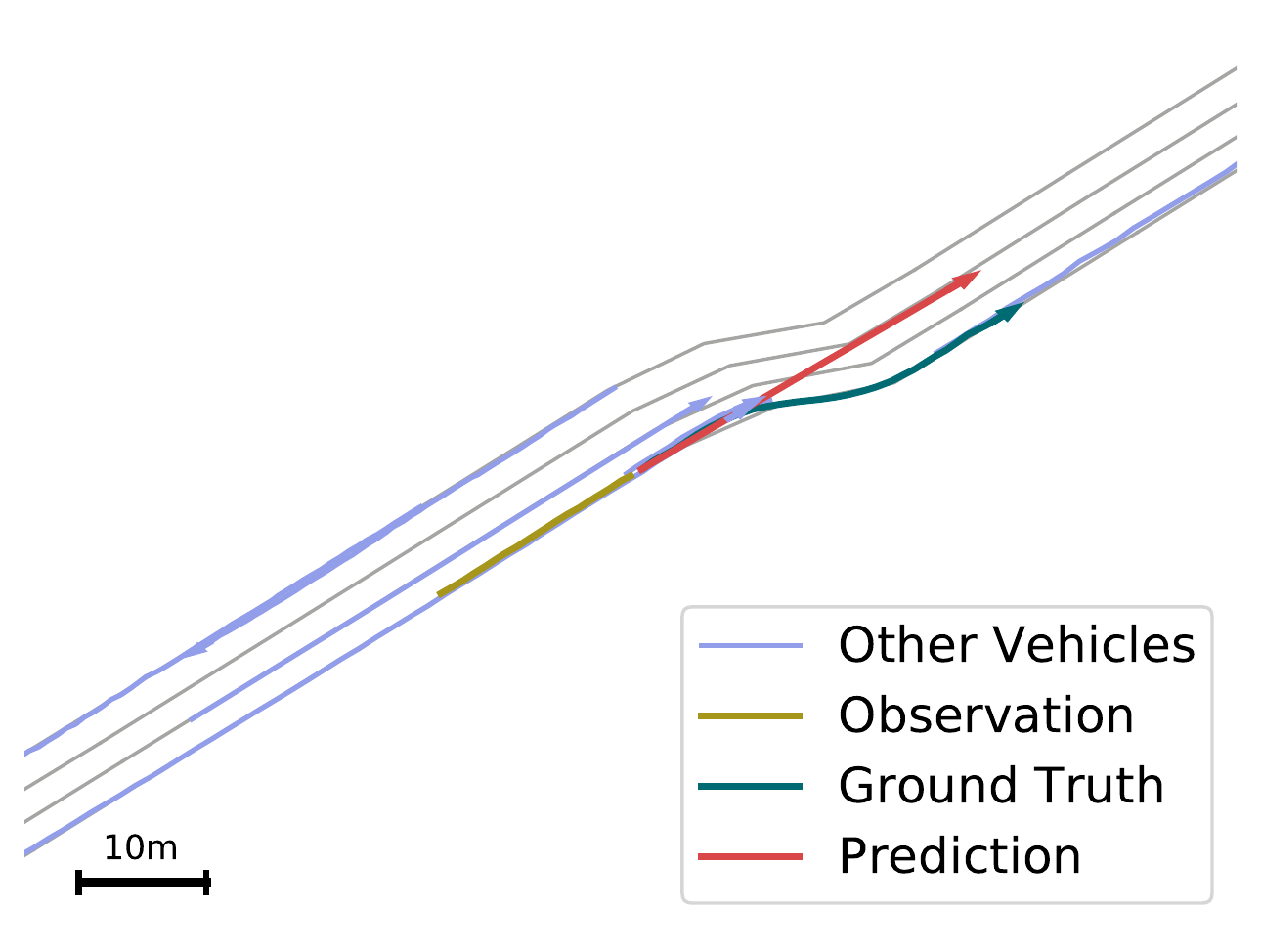}
    \caption{ }
    \label{fig:failure2}
    \end{subfigure}
    \caption{\textbf{Some successful cases of the prediction model.} In (a), the model follows the road and predicts without any off-road. In (b), while the model predicts on-road, it suddenly changes its lane.}
\label{fig:failure}
\end{figure}

\begin{enumerate}
    \item 
We study the ability to transfer the generated scenes to new models, \textit{i.e.,} how models perform on the scenes generated for other models. 
We conduct this experiment by storing the generated scenes for a source model which lead to an off-road prediction, and evaluate the performance of target models on the stored scenes.
\Cref{tab:transferability} shows that the transferred scenes are still difficult cases for other models.

\begin{table}[!t]
\begin{center}
\begin{tabular}{|l|c|c|c|}
\hline
Source models & \multicolumn{3}{c|}{Target models} \\
\hline\hline
& LaneGCN  & DATF & WIMP \\
\hline
LaneGCN  &34 / 100 & 37 / 82 & 20 / 61 \\
DATF   & 11 / 44 & 52 / 100 & 13 / 46 \\
WIMP  & 20 / 63 & 40 / 82  & 36 / 100 \\
\hline
\end{tabular}
\caption{\textbf{Studying the transferability of the generated scenes.} We generate scenes for source model and keep the ones that have off-road prediction by the source model. The target models are evaluated using those scenes. The reported numbers are SOR/HOR values. Numbers are rounded to the nearest integer.}
\label{tab:transferability}
\end{center}
\end{table}

\item We study how models perform with smoothly changing the transformation functions parameters. 
To this end, we smoothly change the transformation parameters for $100$ random scenes and visualize the heatmap of HOR for the generated scenes.
\Cref{fig:cool_exp} demonstrates that models are more vulnerable to larger transformation parameters, \textit{i.e.}, sharper turns. Also, it shows more off-road in the left turns compared with the right ones which could be due to the biases in the dataset~\cite{niedoba2019improving_uber}. A clear improvement is visible in the robust model.

\begin{figure*}[!t]
 \centering
    \centering    \includegraphics[width=0.95\linewidth]{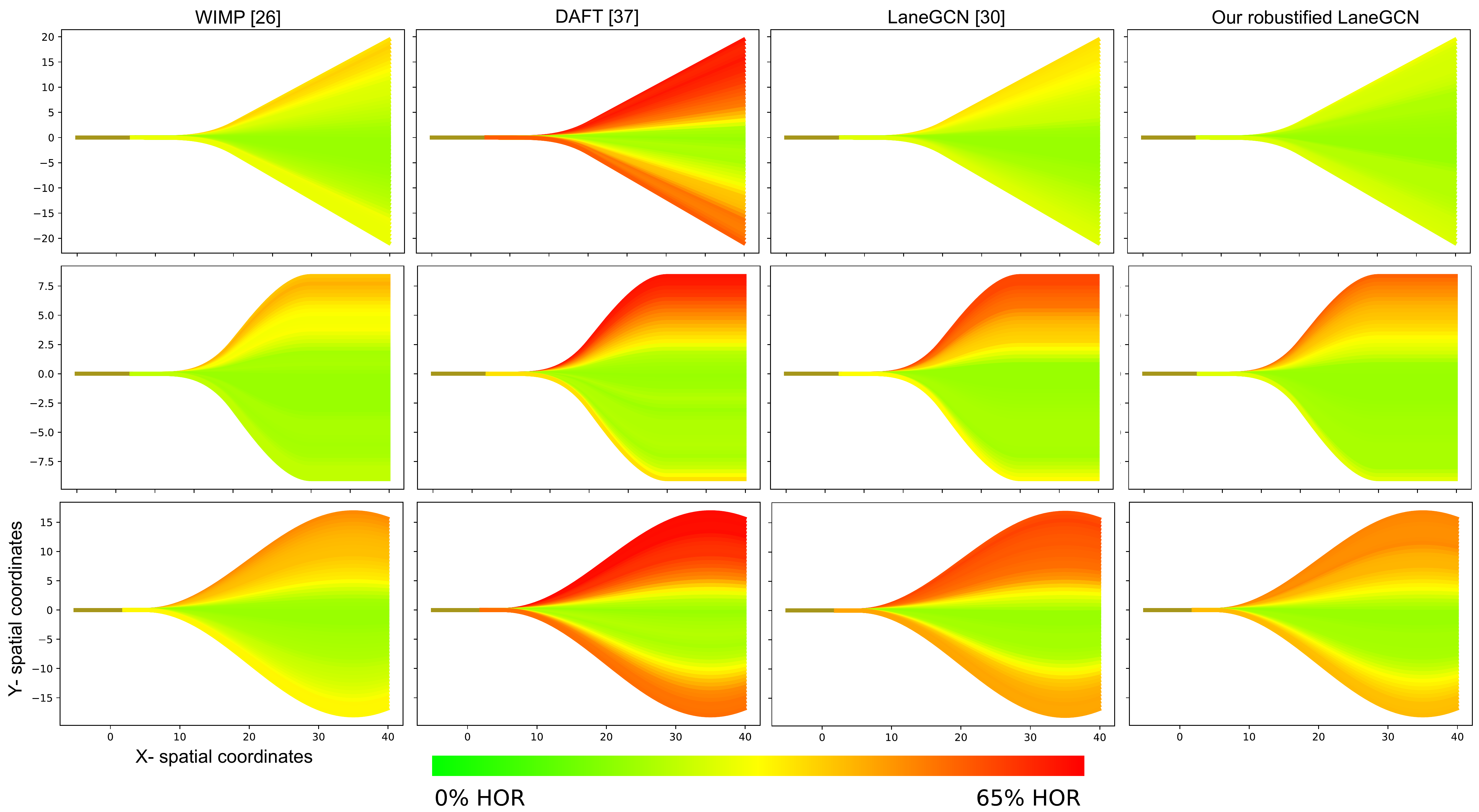}
 \caption{\textbf{The qualitative results of baselines for different transformation functions.}
 The red color indicates more off-road prediction in those scenes and the green indicates higher admissible ones. Usually the models fail in turns with high curvature. We could successfully make the LaneGCN model more robust by fine-tuning.}
 \label{fig:cool_exp}
\end{figure*}

\item Our experiments showed that while the model has almost zero off-road rate in the original scenes, it suffers from over $60\%$ off-road rate in the generated ones. 
In order to hypothesize the causes of this gap, we explored the training data. We observed that in most samples, the history $h$ has enough information about the future trajectory
which reduces the need for the scene reasoning. However, our scene generation approach changes the scene such that $h$ includes almost no information about the future trajectory. 
This essentially makes a situation which requires scene reasoning.
We speculate that this feature is one factor which makes the generated scenes challenging. Note that this does not contradict with the ablations in~\cite{liang2020lanegcn} as their performance measure is accuracy. 
\Cref{fig:speculation1} shows a failure of the model where the prediction is only based on $h$ instead of reasoning over the scene. 
However, the robust model learned to reason over the scene, as shown in \Cref{fig:speculation2}.
While our discussion is an observational hypothesis, we leave further studies for future works. 
\begin{figure}[!t]
 \centering
 \begin{subfigure}[b]{0.49\columnwidth}
    \centering    \includegraphics[width=\linewidth]{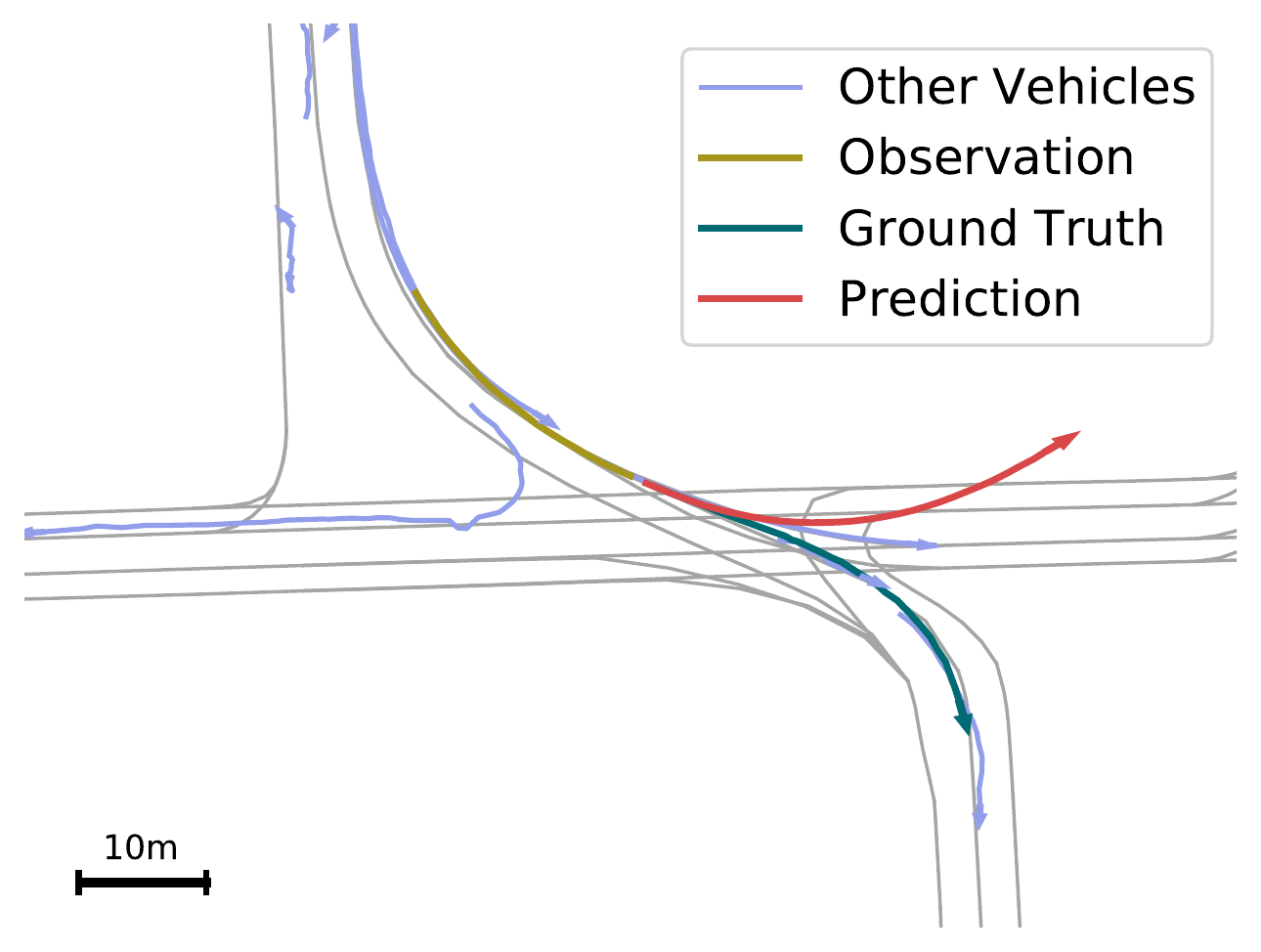}
    \caption{Before robustness}
    \label{fig:speculation1}
    \end{subfigure}
    \hfill
    \begin{subfigure}[b]{0.49\columnwidth}
    \centering    \includegraphics[width=\linewidth]{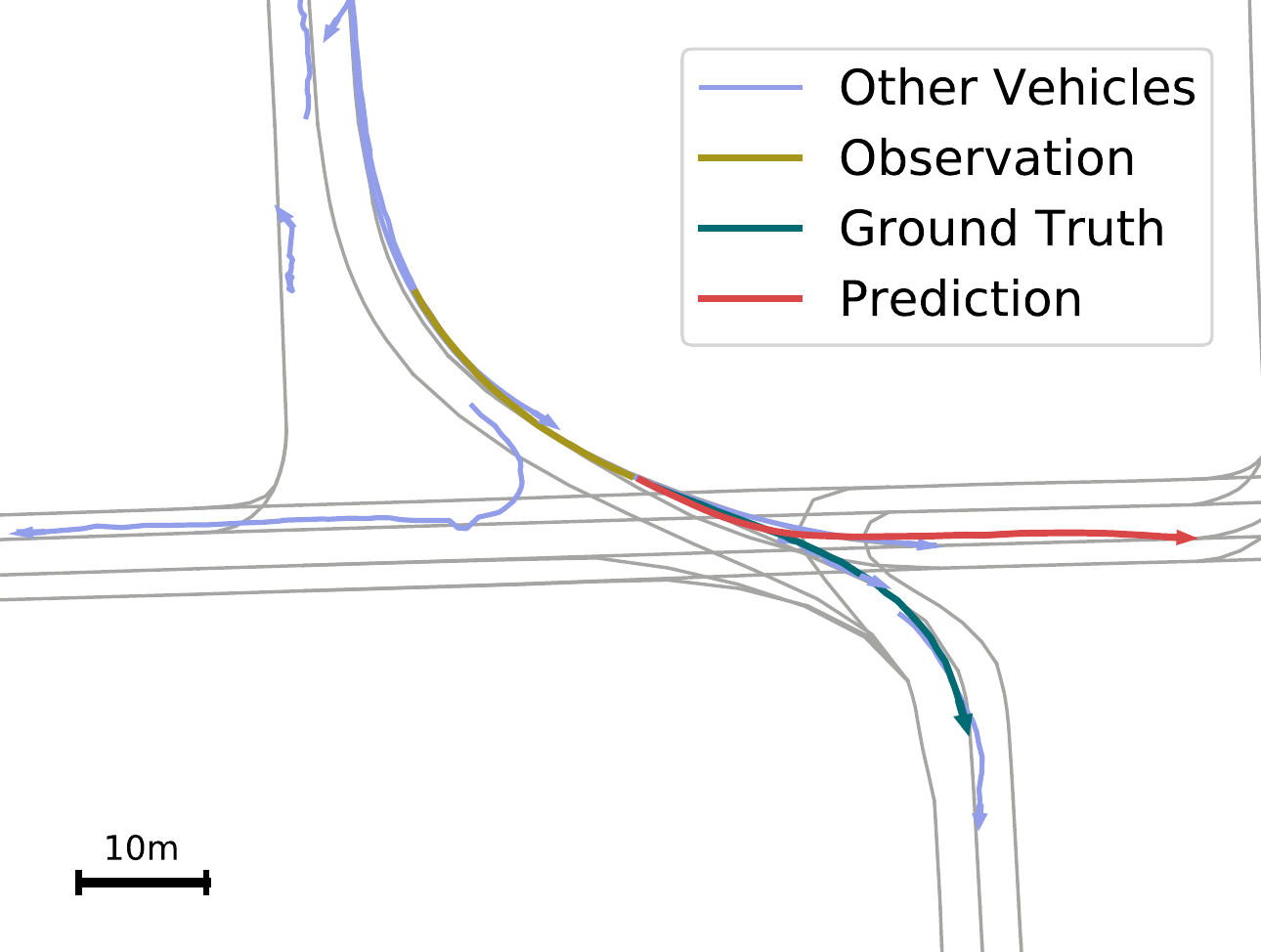}
    \caption{After robustness}
    \label{fig:speculation2}
    \end{subfigure}

 \caption{\textbf{The output of the model before and after the robustness in a sample which requires reasoning over the scene.} We observe that the model before robustness mainly uses $h$ to predict instead of reasoning over the scene. However, after robustness, it reasons more over the scene.}
 \label{fig:speculation}
\end{figure}

\item In some cases, our generated scene could not lead to an off-road prediction.
One such example is depicted in \Cref{fig:failure1}.

\item
\label{sec:lim}
While our method offers a new approach for assessing trajectory prediction models, it has some limitations.
First, our transformation functions are limited, and they cannot cover all real-world cases. We however propose a general methodology that
can be expanded by adding other types of transformations.
To demonstrate it, we add lane merging to the framework, which
causes $14\%$ HOR.
Second, in addition to the off-road criterion, there exist other failure criteria.
For instance, collision with other agents or abnormal behaviors like sudden lane changes. By choosing collision with other agents as criterion, HOR becomes $1.68\%$ in the generated scenes while it is $0.55\%$ in original data. Moreover, \Cref{fig:failure2} shows one scenario in which the predictions of the model are in the drivable area but the sudden lane change is abnormal. 
\end{enumerate}

\section{Conclusion}
In this work, we presented a conditional scene generation method. We showed that several state-of-the-art trajectory prediction models fail in our generated scenes. Notably, they have high off-road rate in their predictions.
Next, leveraging image retrieval techniques, we retrieved real-world
locations which partially resemble the generated scenes and demonstrate their failure in those locations.
We made the model robust against the generated scenes. 
We hope that this framework helps to better evaluate the prediction models which are involved in the autonomous driving systems.

\section{Acknowledgments} 
This project was funded by Honda R\&D Co., Ltd and from the European Union's Horizon 2020 research and innovation programme under the Marie Sklodowska-Curie grant agreement No 754354.
We thank Lorenzo Bertoni and Kirell Benzi for their valuable feedback.

{\small
\bibliographystyle{ieee_fullname}
\bibliography{references}
}

\newpage
\
\newpage

\appendix

\section{Overall algorithm}
\label{sec:overal_alg}
In this section, we demonstrate the overall algorithm for the chosen search method. The pseudo-code of the algorithm for generating a scene is shown in \Cref{alg:alg}. The goal is to generate the scene $S^*$ for a given scenario $x, a, S$ and predictor $g$. The process is called for $k_{max}$ iterations. In each iteration, we start with selecting a transformation function (L. 3). Then, the transformation function generates the corresponding scene (L. 4). After that, the observation trajectory is scaled to ensure feasibility of the scenario (L. 5). Next, the prediction of the model in the new scenario is computed and used to calculate the loss (L. 6, L. 7). The best-achieved loss determines the final generated scene.
\begin{algorithm}[]
\DontPrintSemicolon
\SetAlgoLined
  \KwInput{Sequence $h$, Scene $S$, Predictor $g$, Surrounding vehicles $a$, Transformation set $f$, Number of iterations $k_{max}$} 
  \KwOutput{Generated scene $S^*$}
  Initialize $l^* \gets 1$ \;
  \For{$k= 1 $ \textbf{to} $k_{max}$}{
    Choose a transformation function \;
    $\tilde{S}$ = [$\tilde{s}$] where $\tilde{s}$ $\gets$ \Cref{eq:general_transf} \;
    Obtain $\tilde{h}$, $\tilde{a}$ from phys constraints \Cref{sec:nat_scen} \;
    $\tilde{z} = g(\tilde{h}, \tilde{S}, \tilde{a})$ \;
    Calculate $l$ using \Cref{eq:opt} \;
    \If{$l<l^*$}{
    ${S^*} = \tilde{S}$ \;
    }
   }
\caption{Scene search method}
\label{alg:alg}
\end{algorithm}

\section{Additional qualitative results}
\label{sec:appendix:qual}
\begin{enumerate}

    \item \textbf{Real-world retrieval images.}
We show more real-world examples for both cases where the trajectory prediction model fails and succeeds in \Cref{fig:real-world-supp}.

\begin{figure*}[!t]
  \centering
  \begin{subfigure}[b]{0.327\linewidth}
    \centering\includegraphics[width=\linewidth]{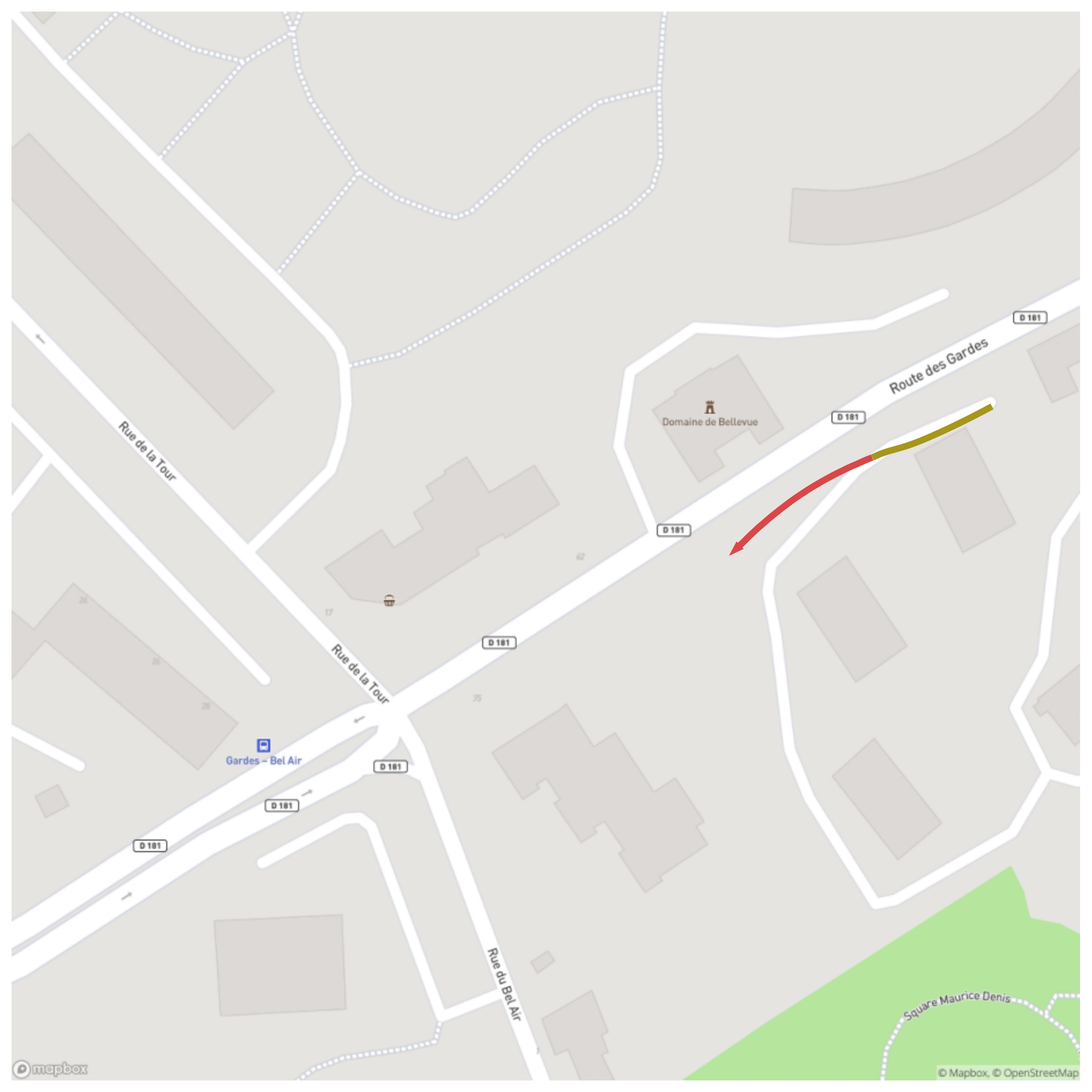}
    \caption{Paris 
    \href{https://www.openstreetmap.org/\#map=19/48.81605/2.22530}{location}}
  \end{subfigure}
  \hfill
  \begin{subfigure}[b]{0.327\linewidth}
    \centering\includegraphics[width=\linewidth]{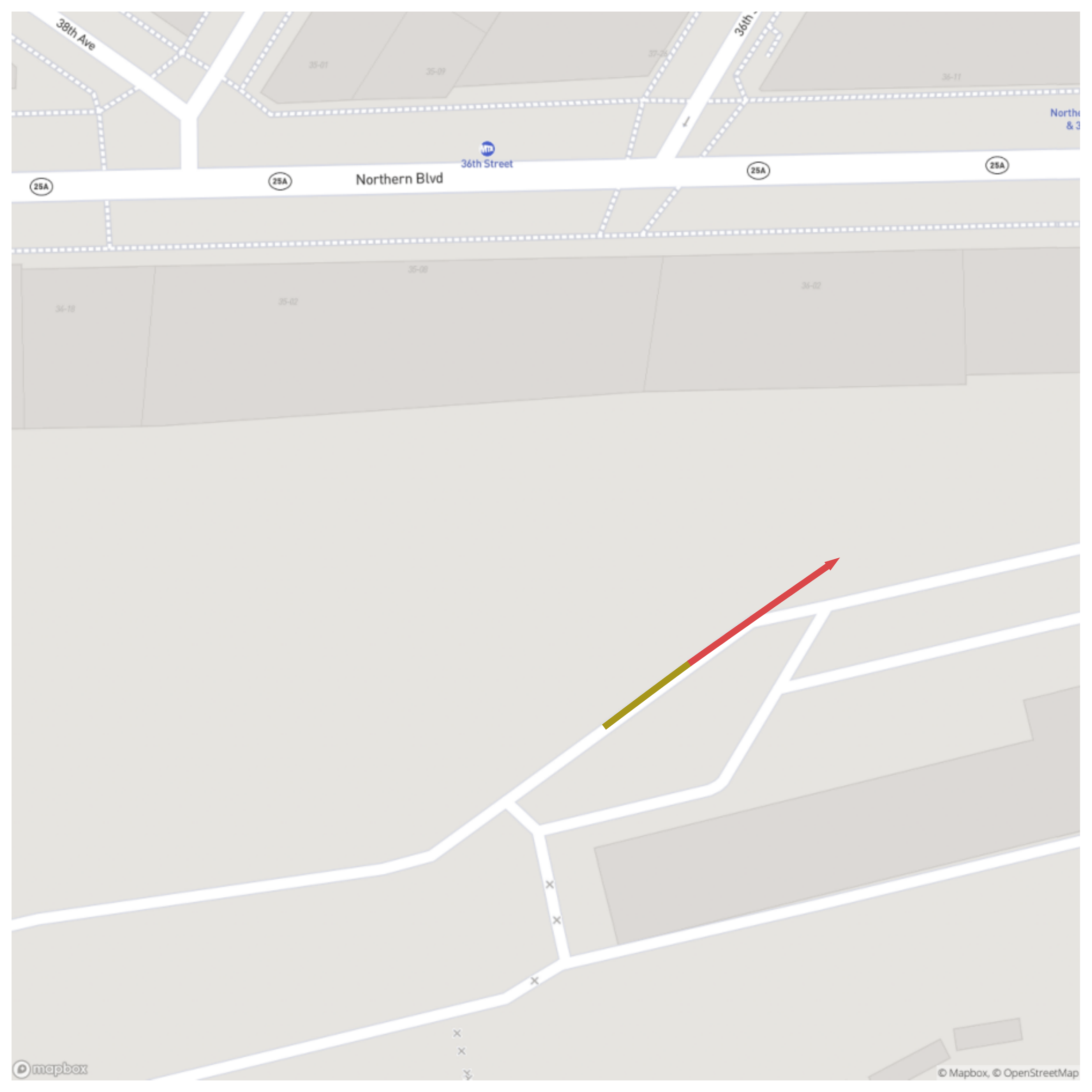}
    \caption{New York \href{https://www.openstreetmap.org/\#map=19/40.75142/-73.92815}{location}}
  \end{subfigure}
  \hfill
    \begin{subfigure}[b]{0.327\linewidth}
    \centering\includegraphics[width=\linewidth]{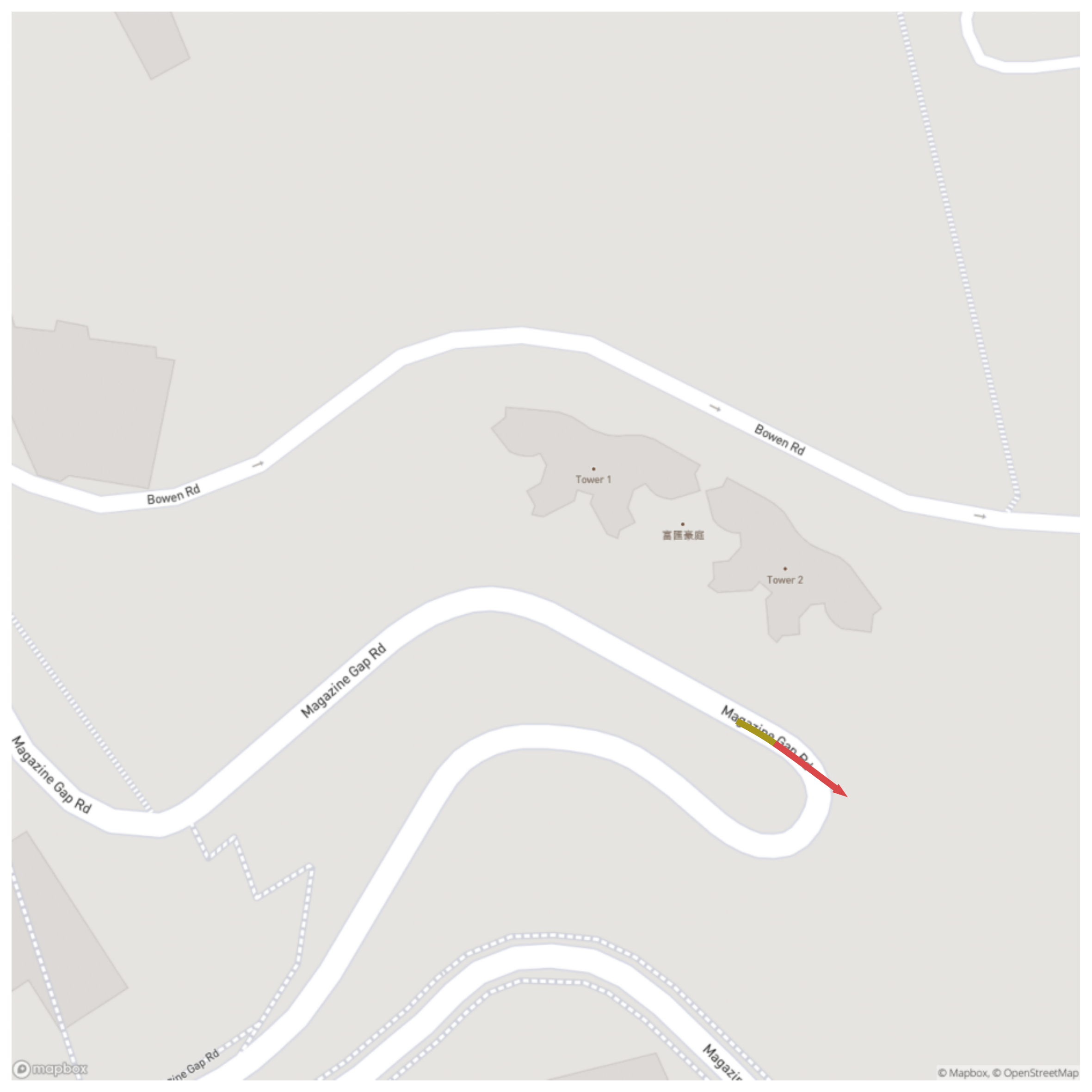}
    \caption{Hong Kong \href{https://www.openstreetmap.org/\#map=19/22.2736954/114.1589233}{location}}
  \end{subfigure}
  \\
  \begin{subfigure}[b]{0.327\linewidth}
    \centering\includegraphics[width=\linewidth]{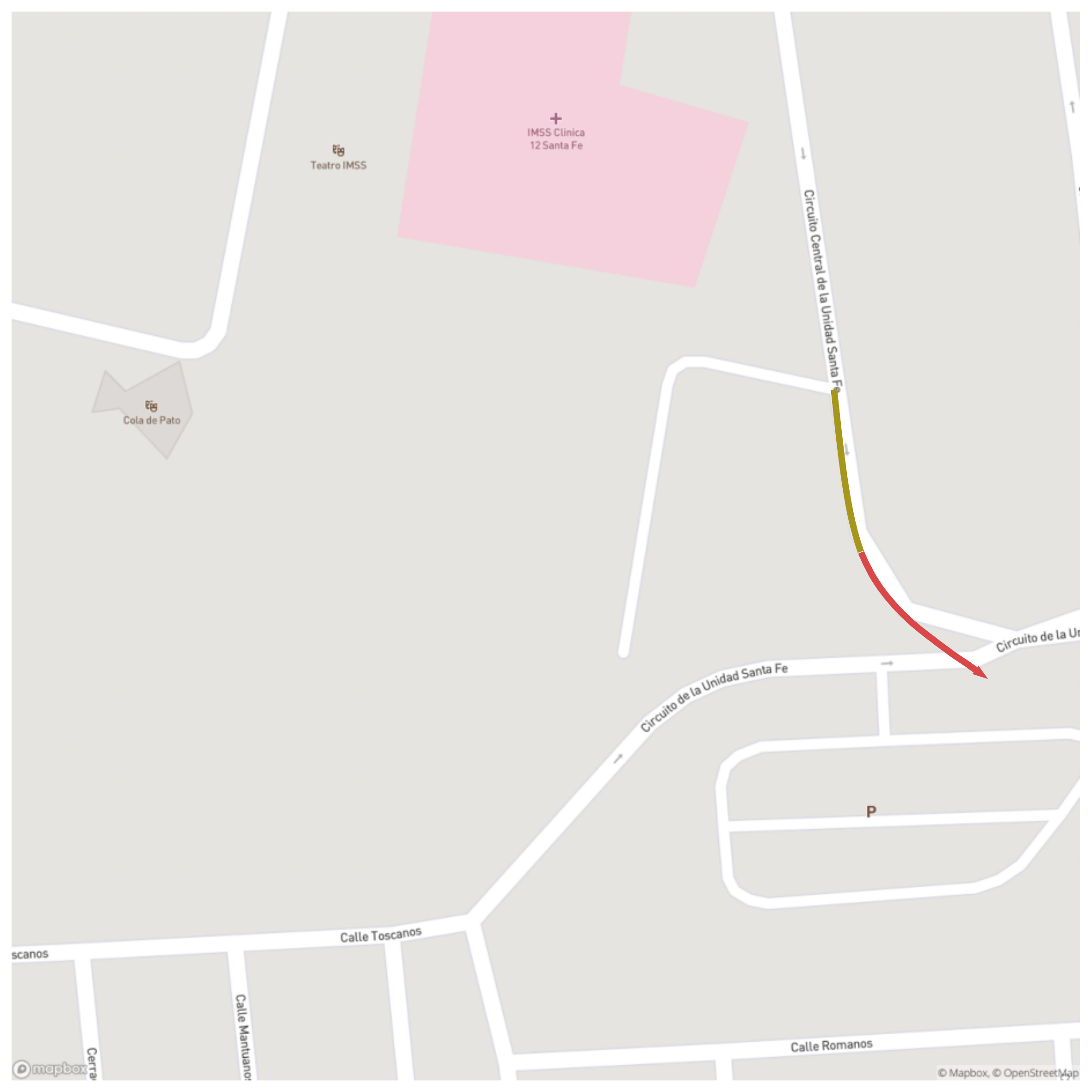}
    \caption{New Mexico \href{https://www.openstreetmap.org/\#map=19/19.38772/-99.20424}{location}}
  \end{subfigure}
  \hfill
  \begin{subfigure}[b]{0.327\linewidth}
    \centering\includegraphics[width=\linewidth]{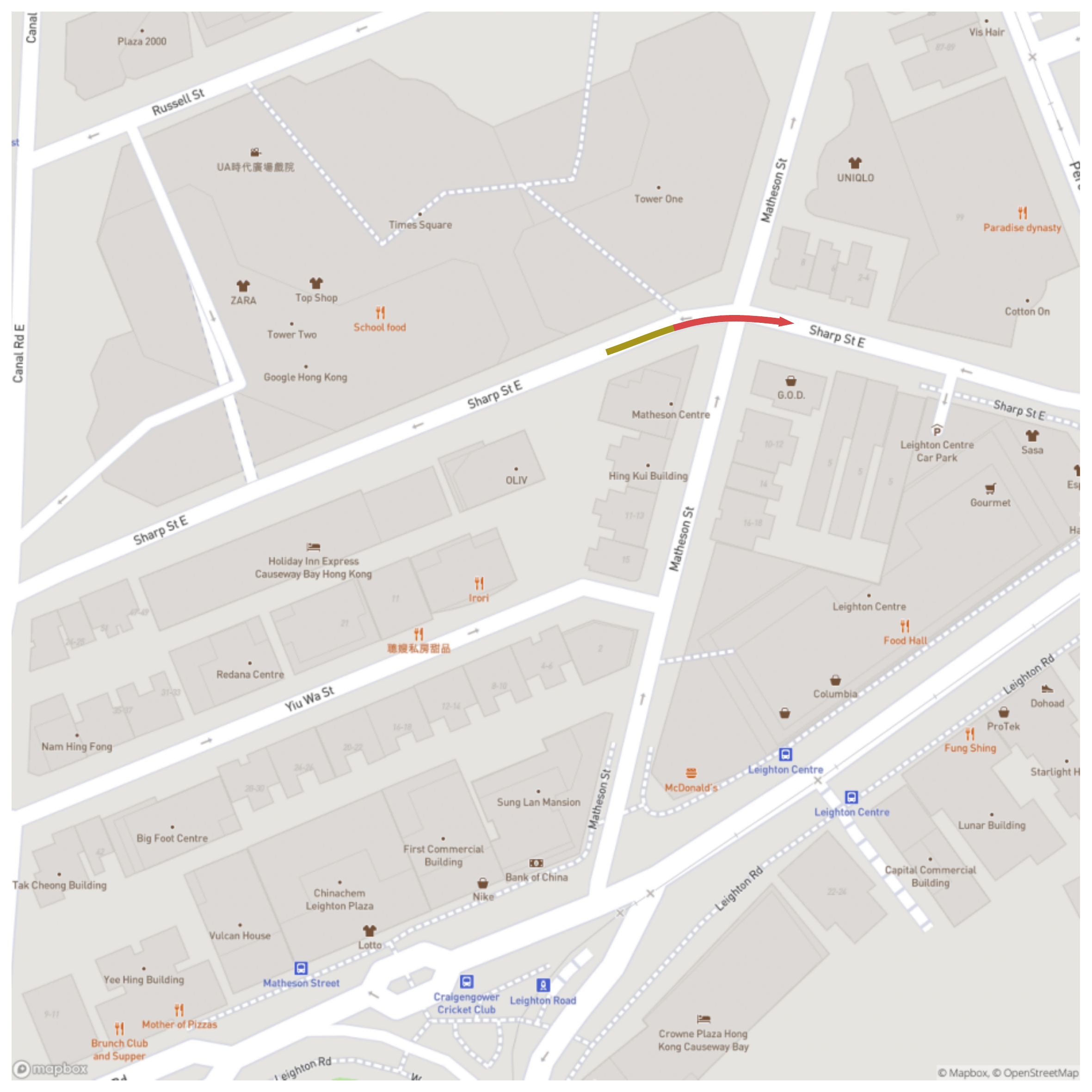}
    \caption{Hong Kong \href{https://www.openstreetmap.org/\#map=19/22.278162/114.182714}{location}}
  \end{subfigure}
  \hfill
    \begin{subfigure}[b]{0.327\linewidth}
    \centering\includegraphics[width=\linewidth]{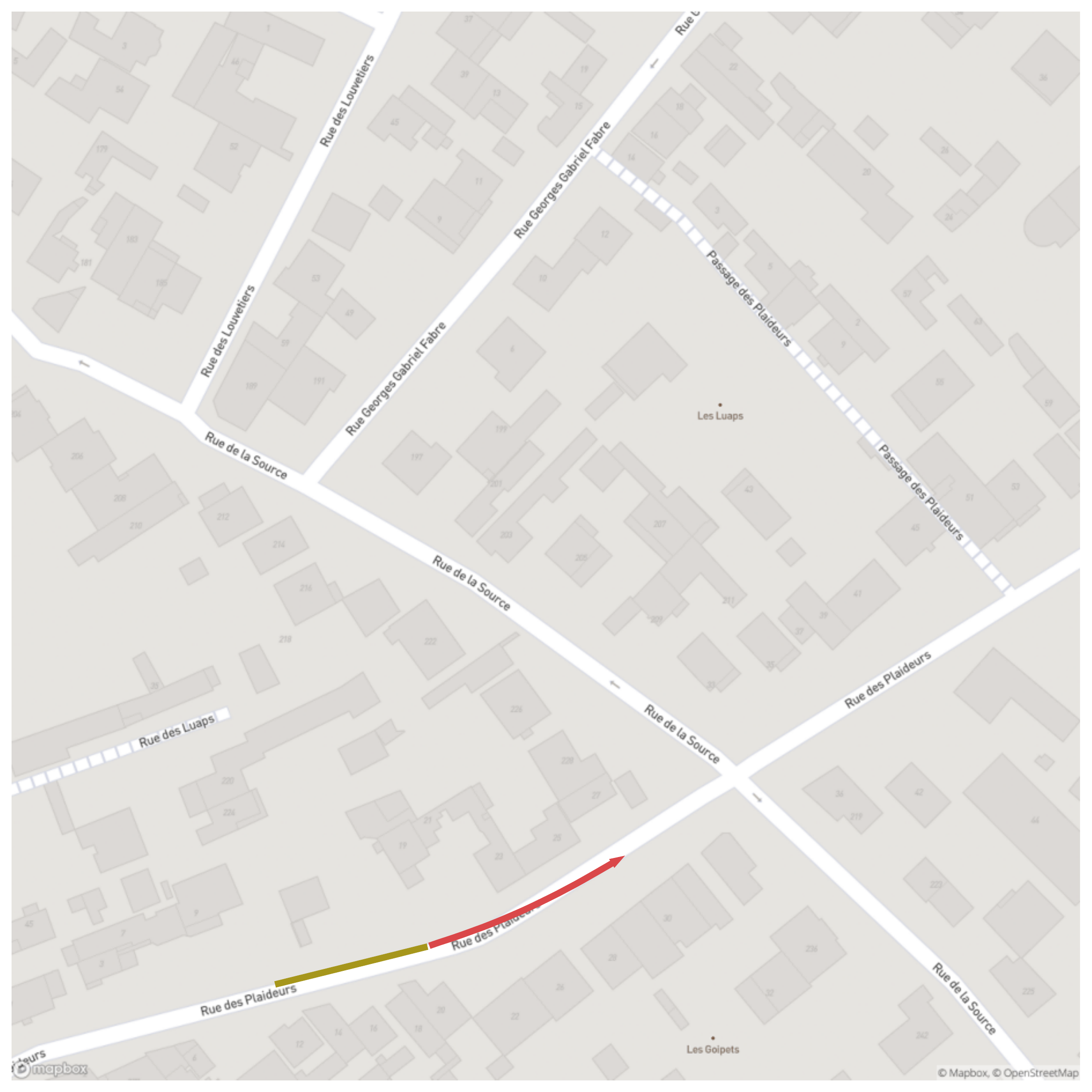}
    \caption{Paris \href{https://www.openstreetmap.org/\#map=19/48.878975/2.212208}{location}}
    \end{subfigure}
  \\
  \begin{subfigure}[b]{0.327\linewidth}
    \centering\includegraphics[width=\linewidth]{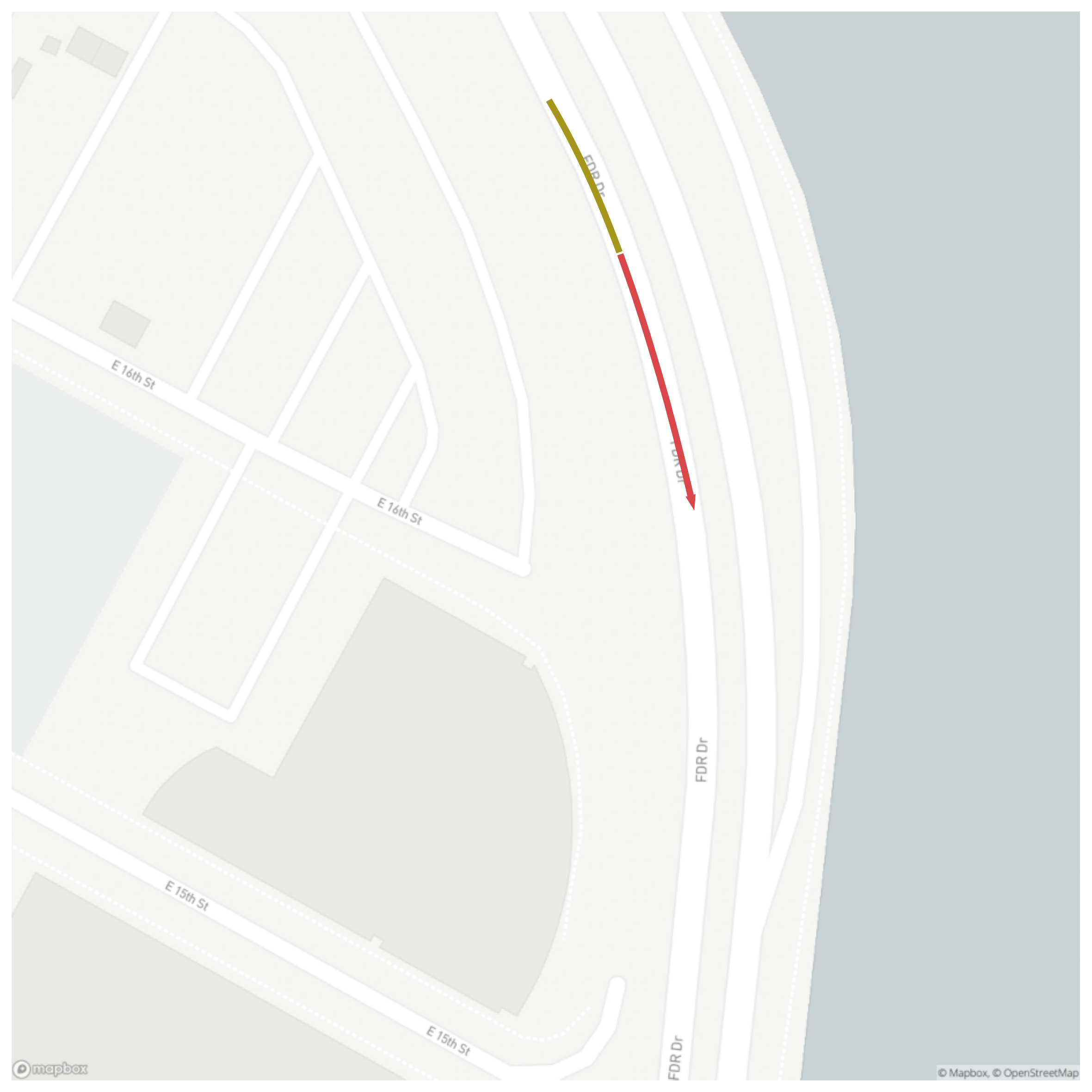}
    \caption{New York \href{https://www.openstreetmap.org/\#map=19/40.72878/-73.97212}{location}}
  \end{subfigure}
  \hfill
  \begin{subfigure}[b]{0.327\linewidth}
    \centering\includegraphics[width=\linewidth]{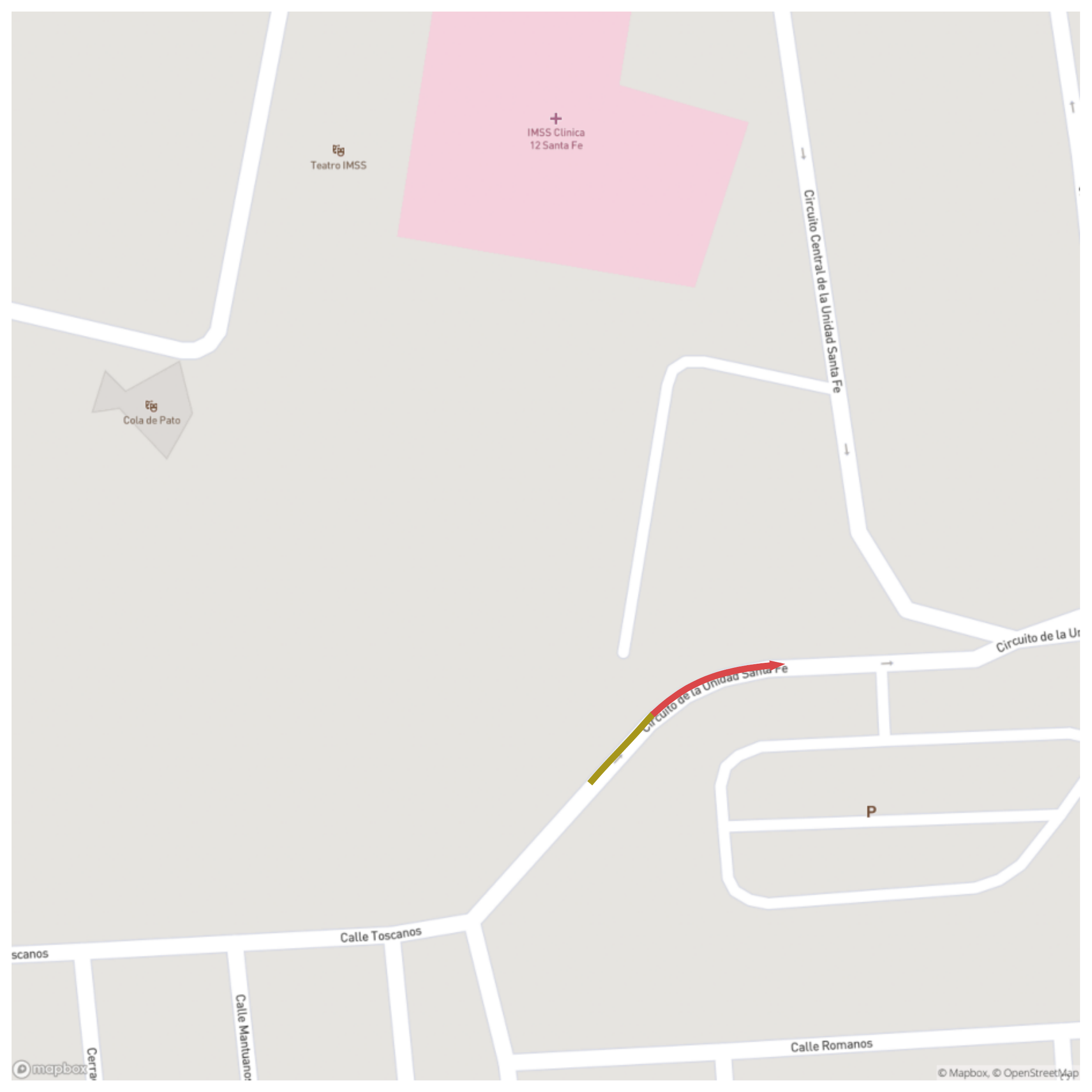}
    \caption{New Mexico \href{https://www.openstreetmap.org/\#map=19/19.38772/-99.20424}{location}}
  \end{subfigure}
  \hfill
    \begin{subfigure}[b]{0.327\linewidth}
    \centering\includegraphics[width=\linewidth]{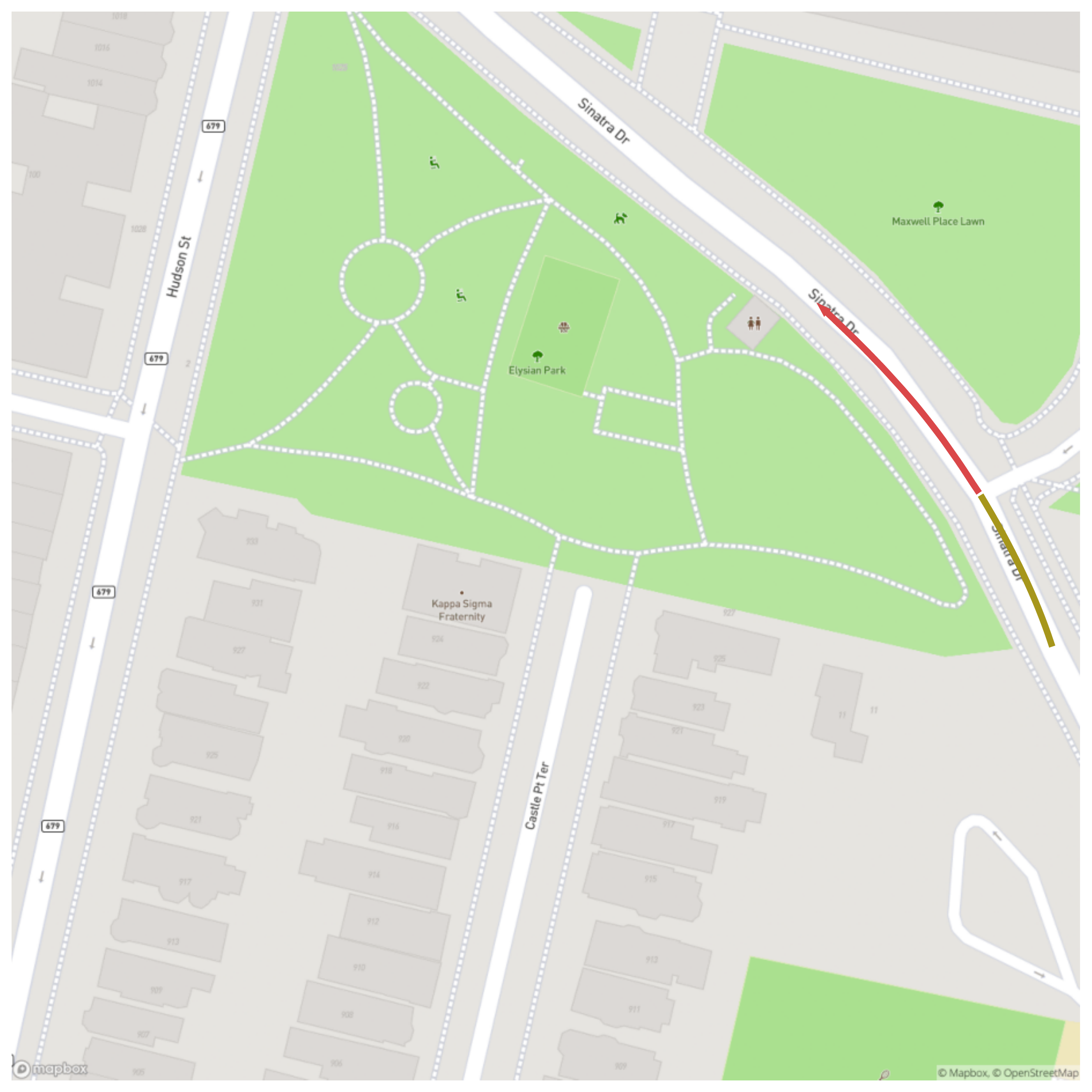}
    \caption{New York \href{https://www.openstreetmap.org/\#map=19/40.748401/-74.024646}{location}}
  \end{subfigure}
  \caption{
  \textbf{Retrieving real-world places using our real-world retrieval algorithm.} We observe that the model fails in Paris (a), New York (b), Hong Kong (c) and New Mexico (d). The model also successfully predicts in the drivable area in the remaining figures.}
 \label{fig:real-world-supp}
\end{figure*}

    \item \textbf{More generated scenes.}
\Cref{fig:all-supp} provides more visualizations for the performance of the baselines in our generated scenes.

\begin{figure*}[!t]
  \centering
  \begin{subfigure}[b]{0.327\linewidth}
    \centering\includegraphics[width=\linewidth]{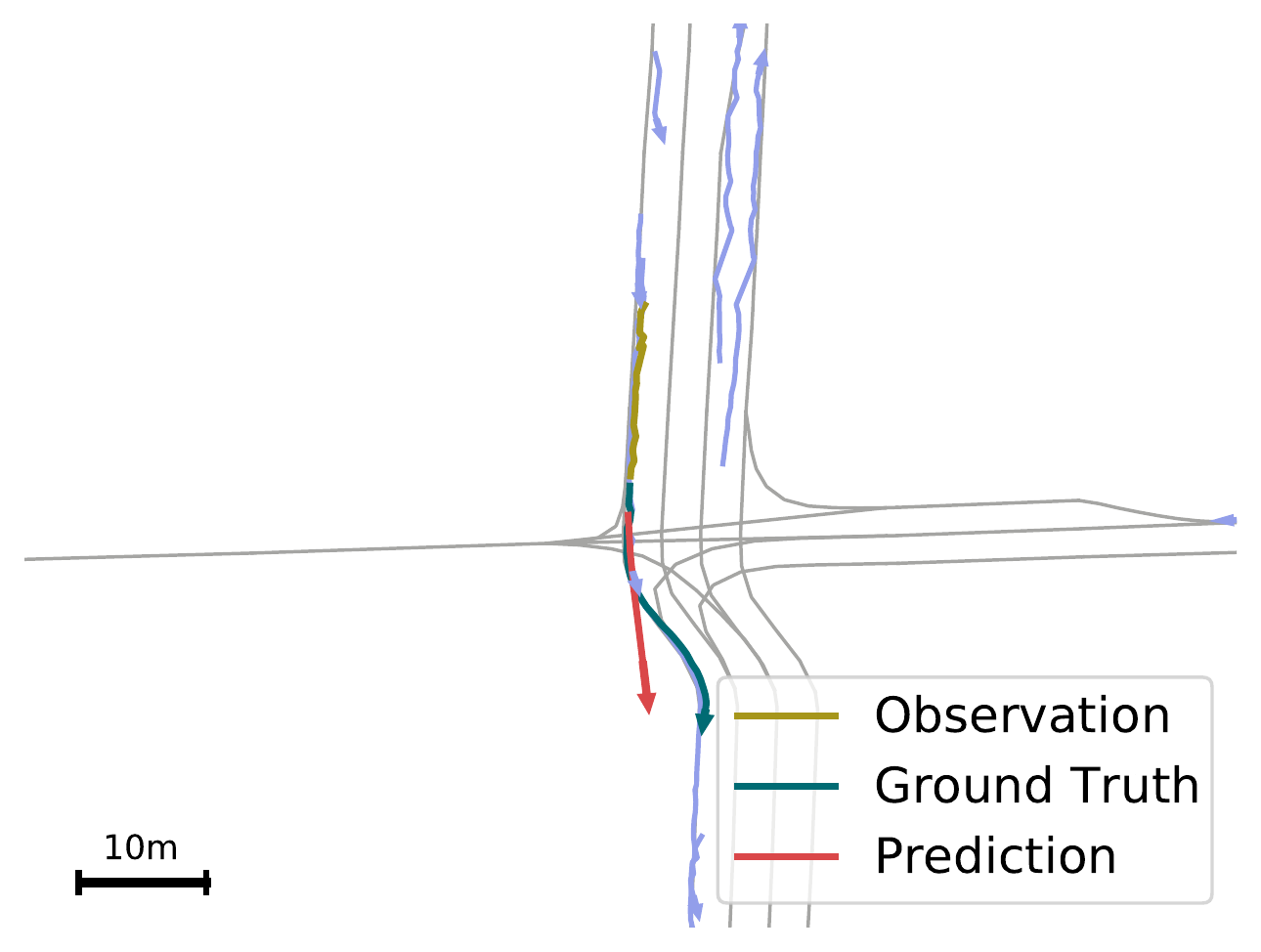}
    \caption{~DATF}
  \end{subfigure}
  \hfill
  \begin{subfigure}[b]{0.327\linewidth}
    \centering\includegraphics[width=\linewidth]{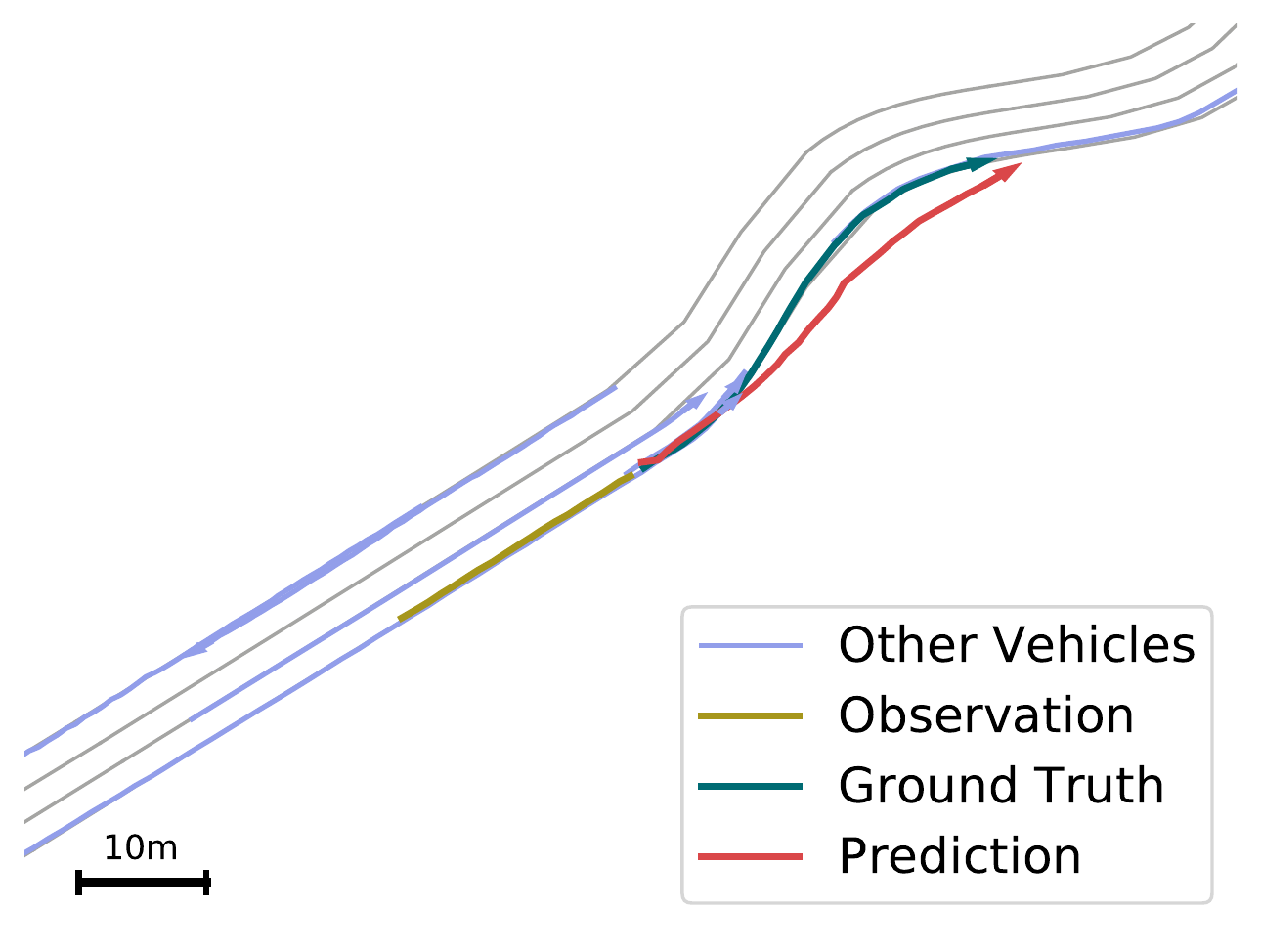}
    \caption{~WIMP}
  \end{subfigure}
  \hfill
  \begin{subfigure}[b]{0.327\linewidth}
    \centering\includegraphics[width=\linewidth]{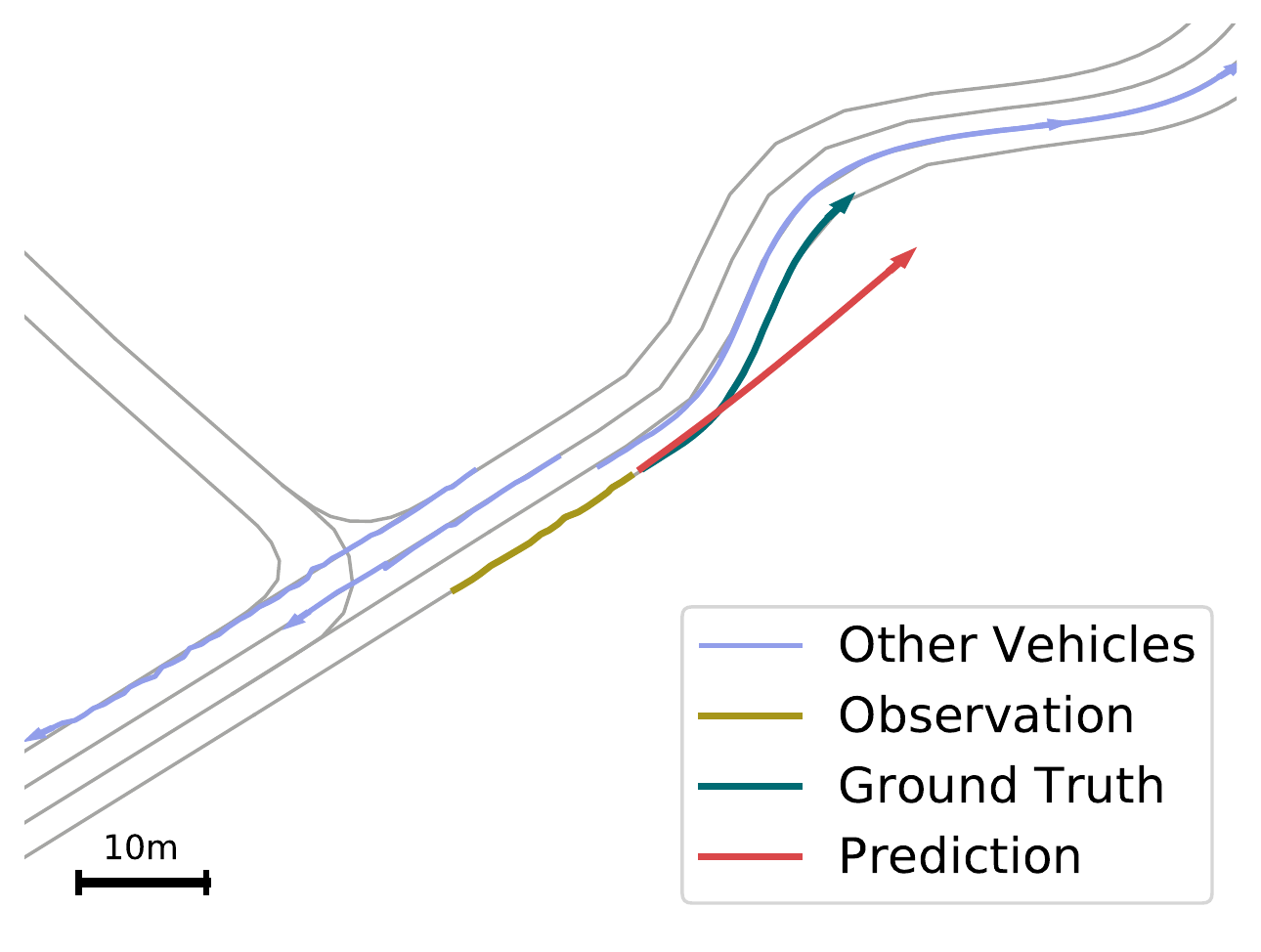}
    \caption{~LaneGCN}
  \end{subfigure} \\
  \begin{subfigure}[b]{0.327\linewidth}
    \centering\includegraphics[width=\linewidth]{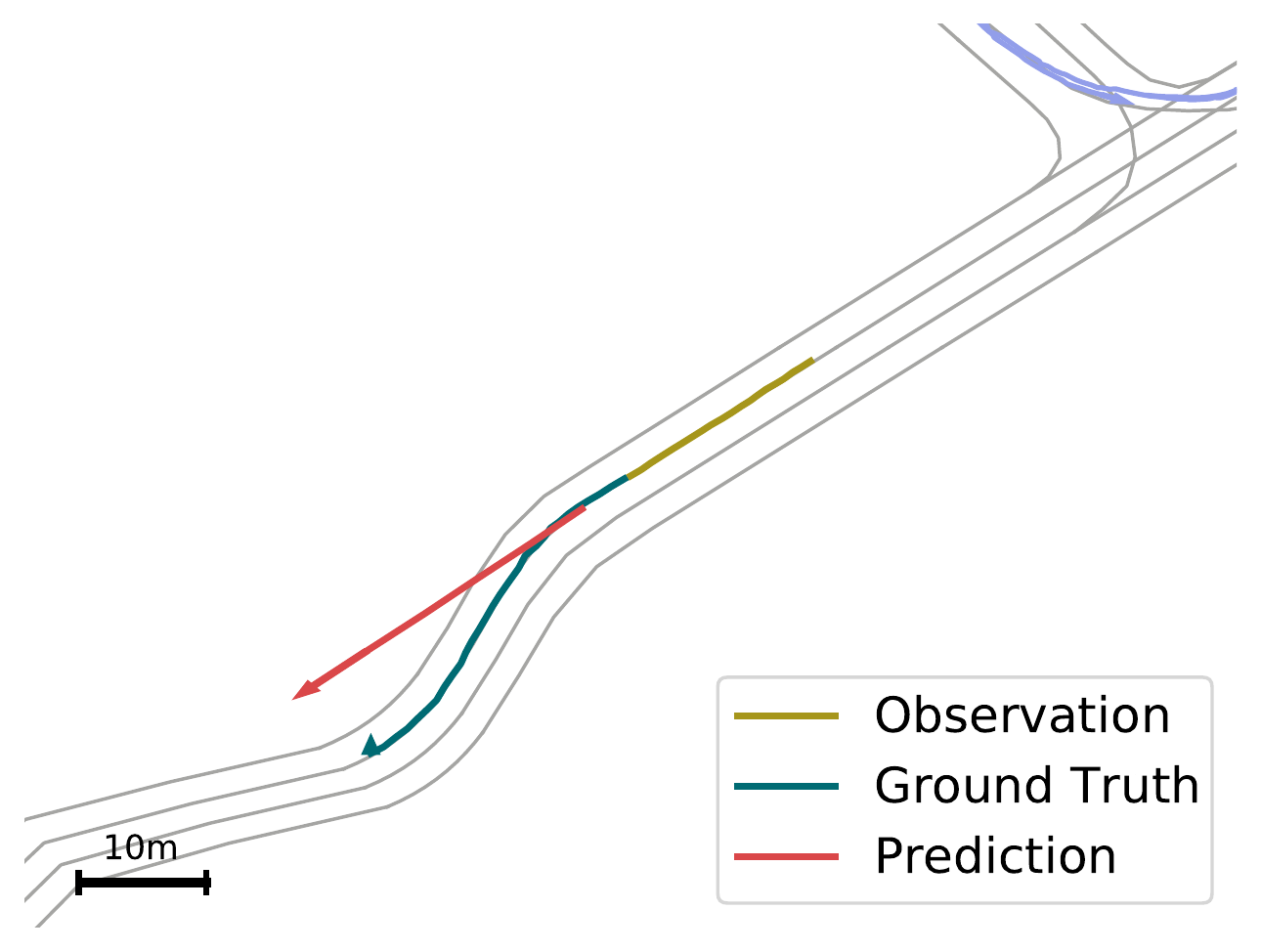}
    \caption{~DATF}
  \end{subfigure}
  \hfill
  \begin{subfigure}[b]{0.327\linewidth}
    \centering\includegraphics[width=\linewidth]{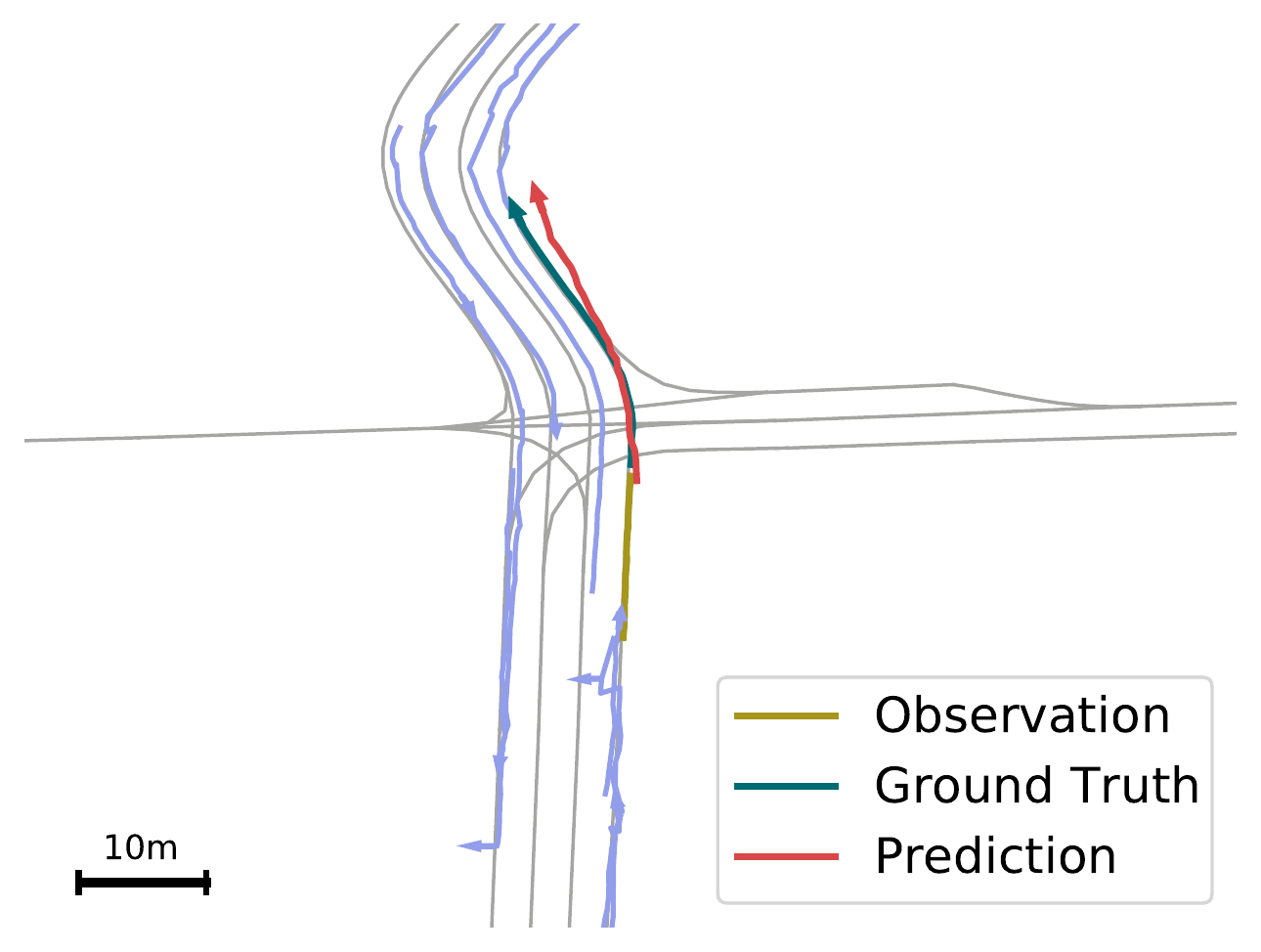}
    \caption{~WIMP}
  \end{subfigure}
  \hfill
  \begin{subfigure}[b]{0.327\linewidth}
    \centering\includegraphics[width=\linewidth]{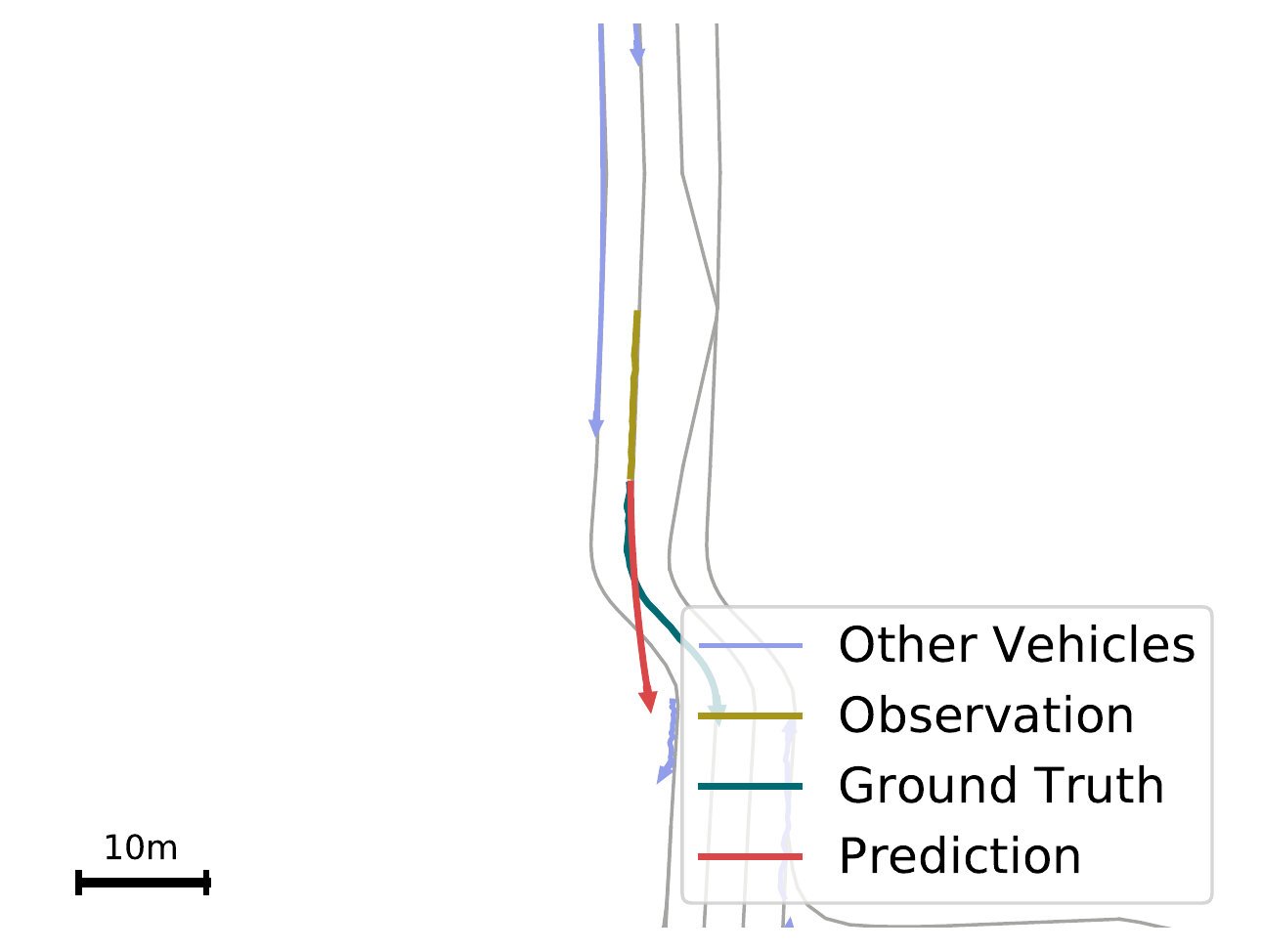}
    \caption{~LaneGCN}
  \end{subfigure} \\
  \begin{subfigure}[b]{0.327\linewidth}
    \centering\includegraphics[width=\linewidth]{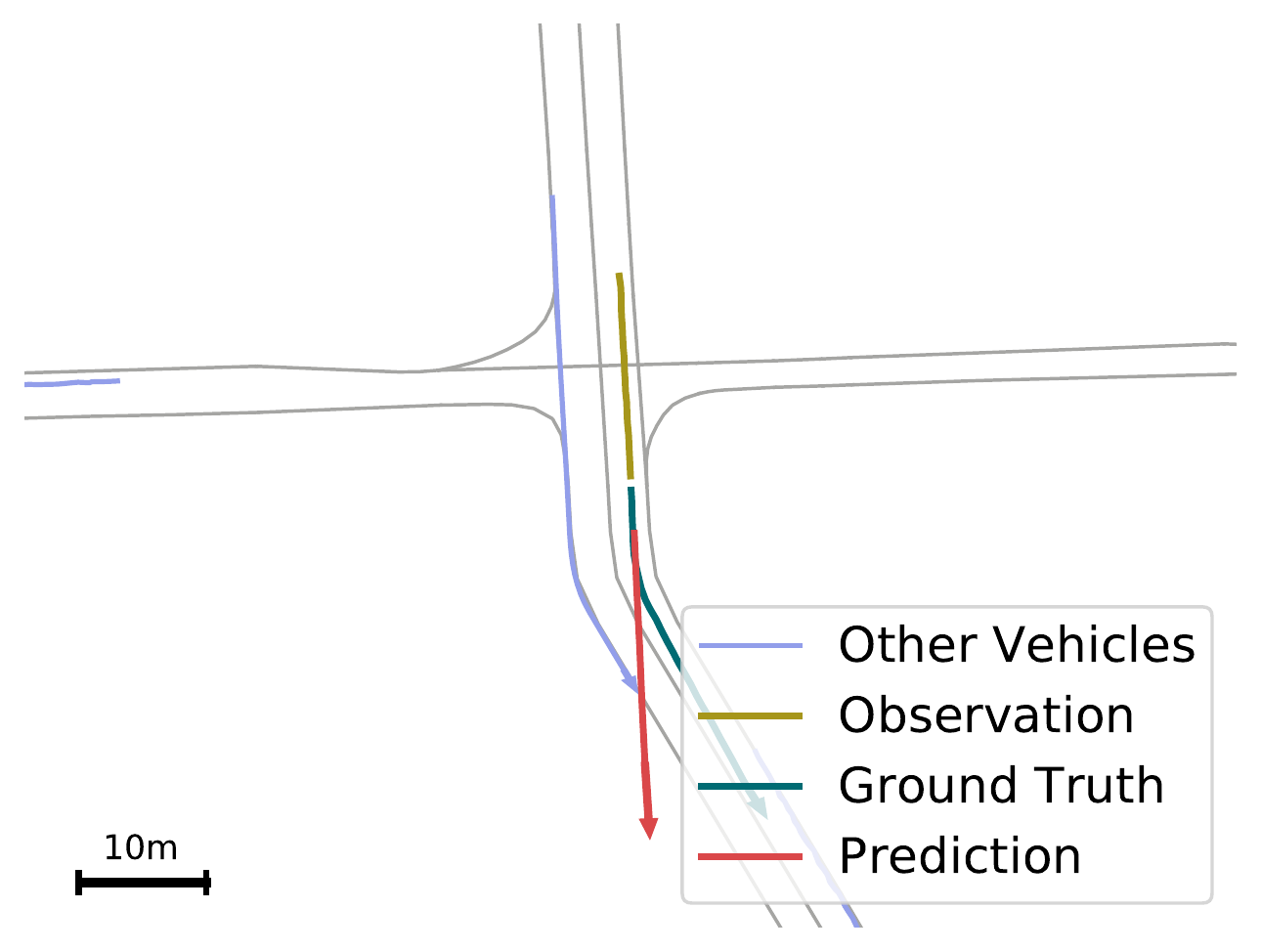}
    \caption{~DATF}
  \end{subfigure}
  \hfill
  \begin{subfigure}[b]{0.327\linewidth}
    \centering\includegraphics[width=\linewidth]{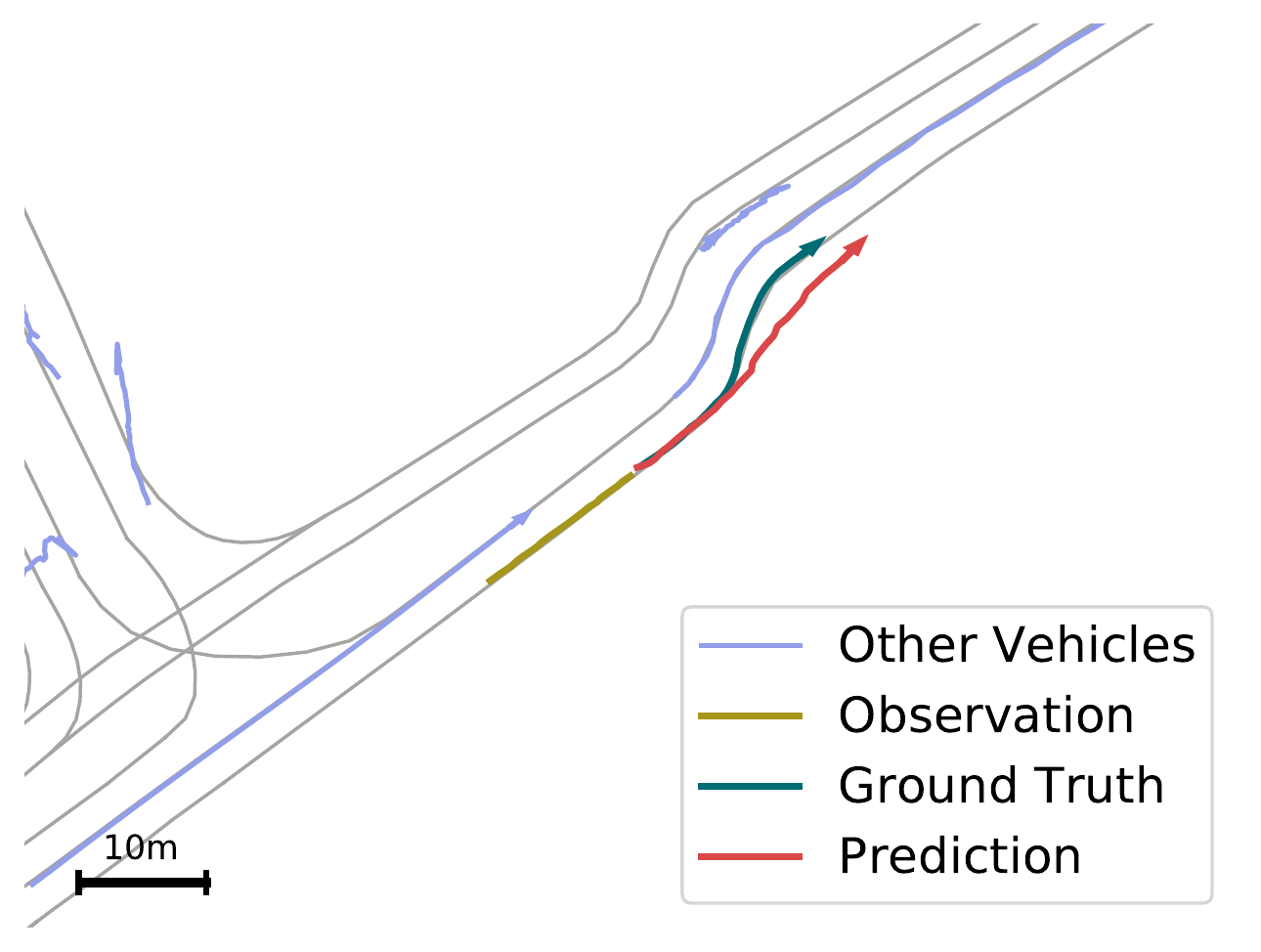}
    \caption{~WIMP}
  \end{subfigure}
  \hfill
  \begin{subfigure}[b]{0.327\linewidth}
    \centering\includegraphics[width=\linewidth]{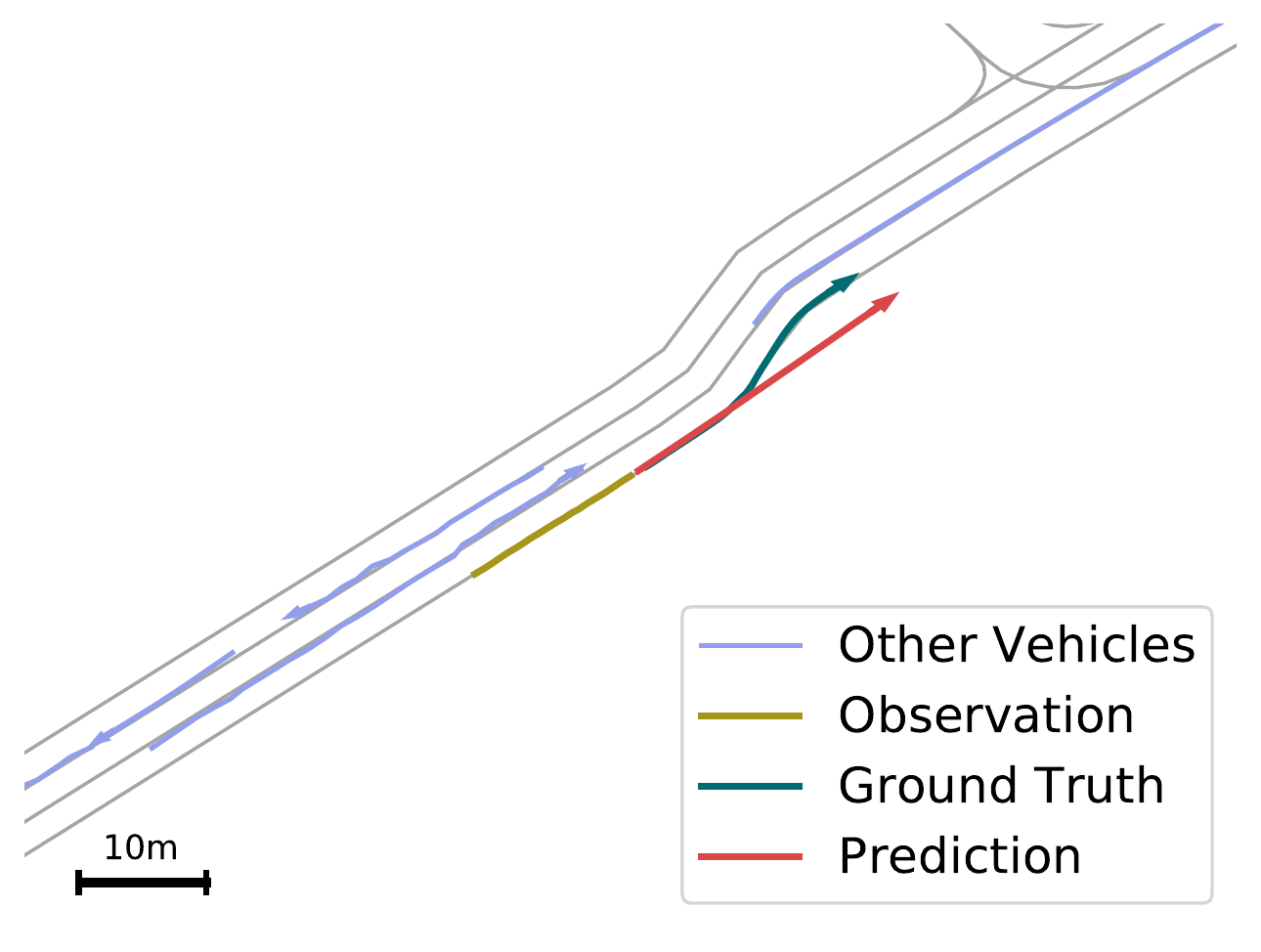}
    \caption{~LaneGCN}
  \end{subfigure}
  \caption{
  \textbf{The predictions of different models in some generated scenes.} All models are challenged by the generated scenes and failed in predicting in the drivable area.
}
 \label{fig:all-supp}
\end{figure*}

    \item \textbf{Noise in the drivable area map.}
The models predict near perfect in the original dataset with HOR of less than $1\%$. Our exploration shows that most of the $1\%$ failed cases are due to the annotation noise in the drivable area maps of the dataset and the models are almost error-free with respect to the scene. Some figures are provided in \Cref{fig:noise-issue}.

\begin{figure*}[!t]
    \centering
    \begin{subfigure}[b]{0.327\linewidth}
    \includegraphics[width=\linewidth]{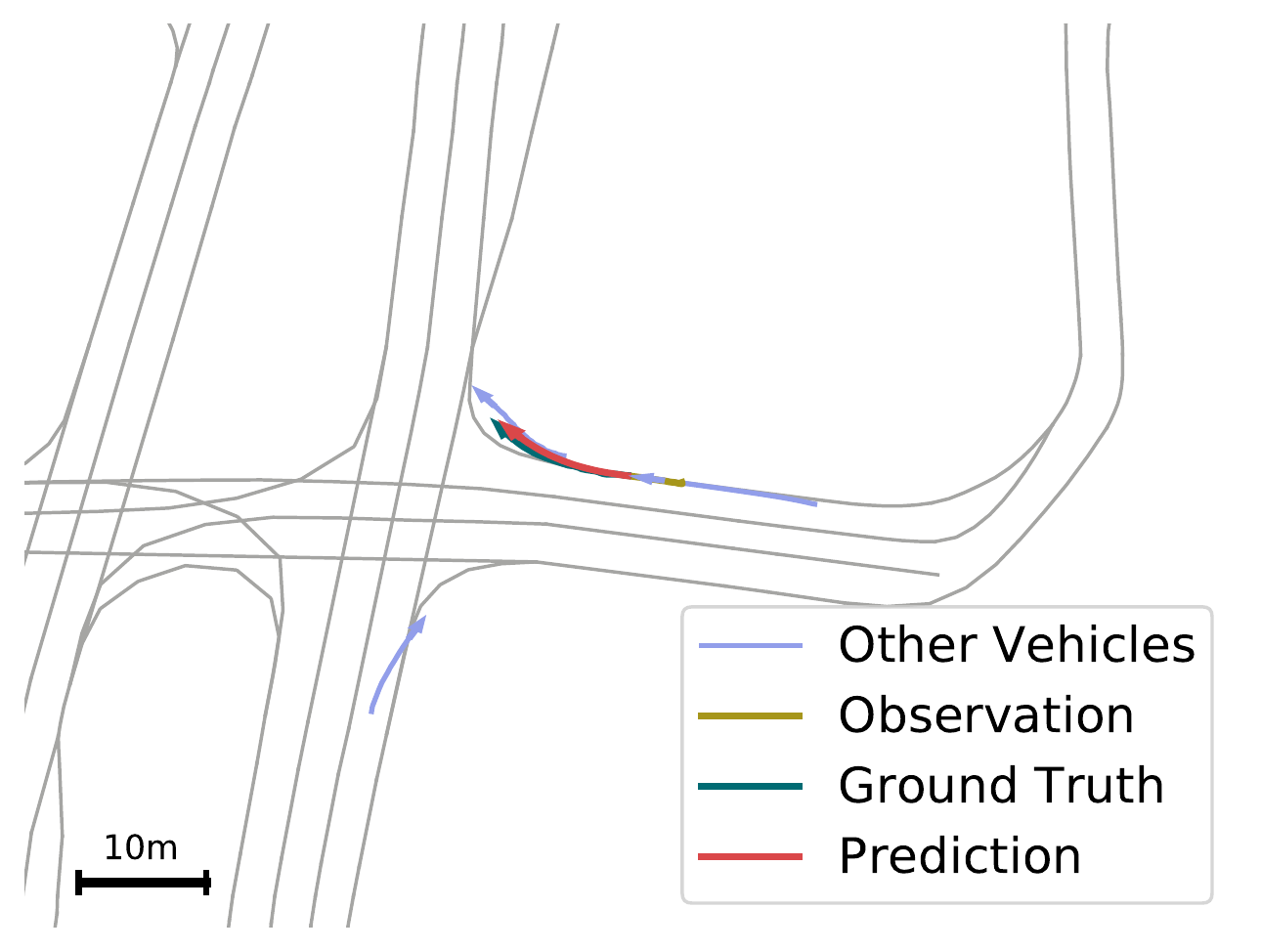}
    \caption{ }
    \end{subfigure}
    \hfill
    \begin{subfigure}[b]{0.327\linewidth}
    \includegraphics[width=\linewidth]{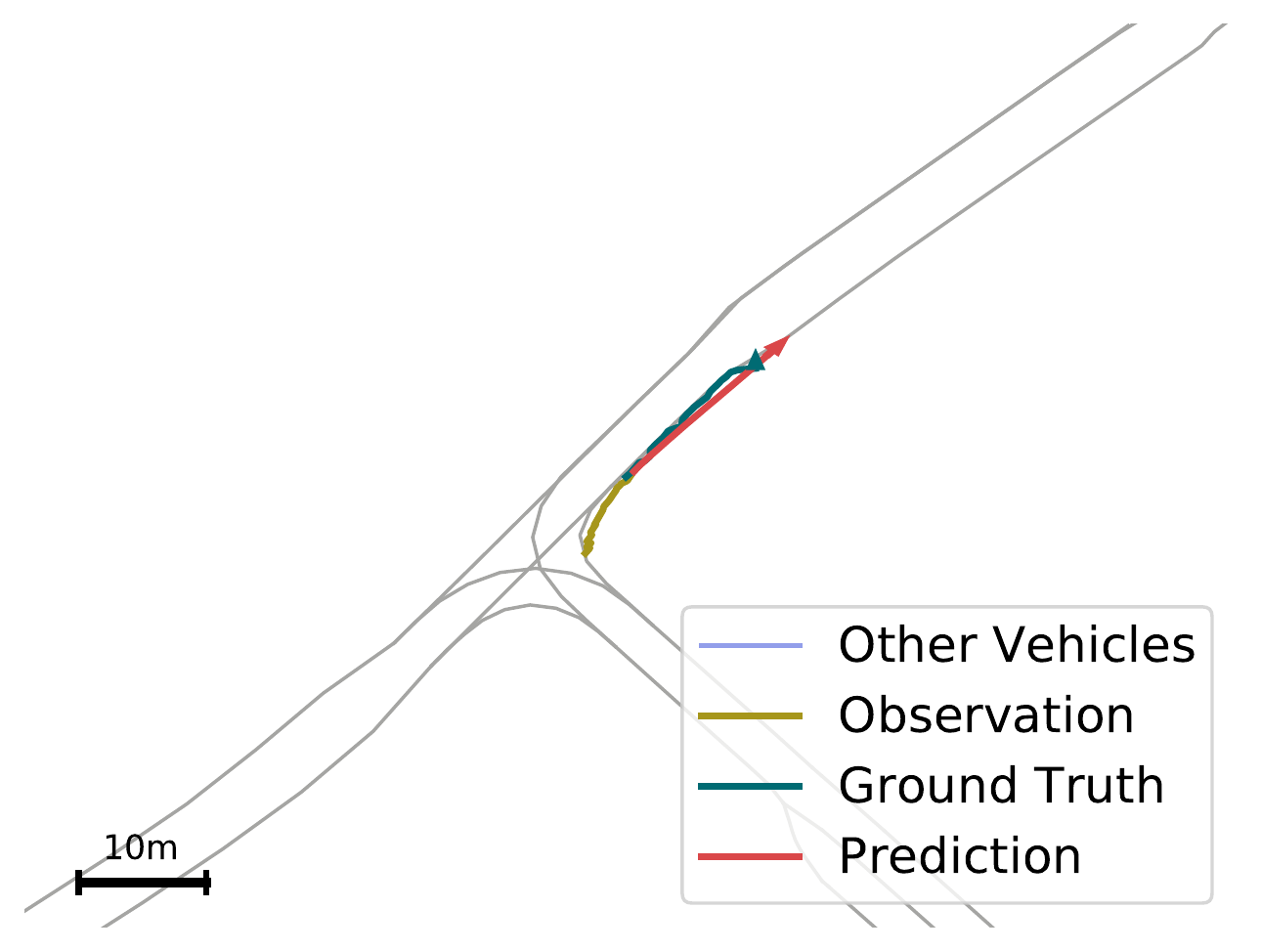}
    \caption{ }
    \end{subfigure}
    \hfill
    \begin{subfigure}[b]{0.327\linewidth}
    \includegraphics[width=\linewidth]{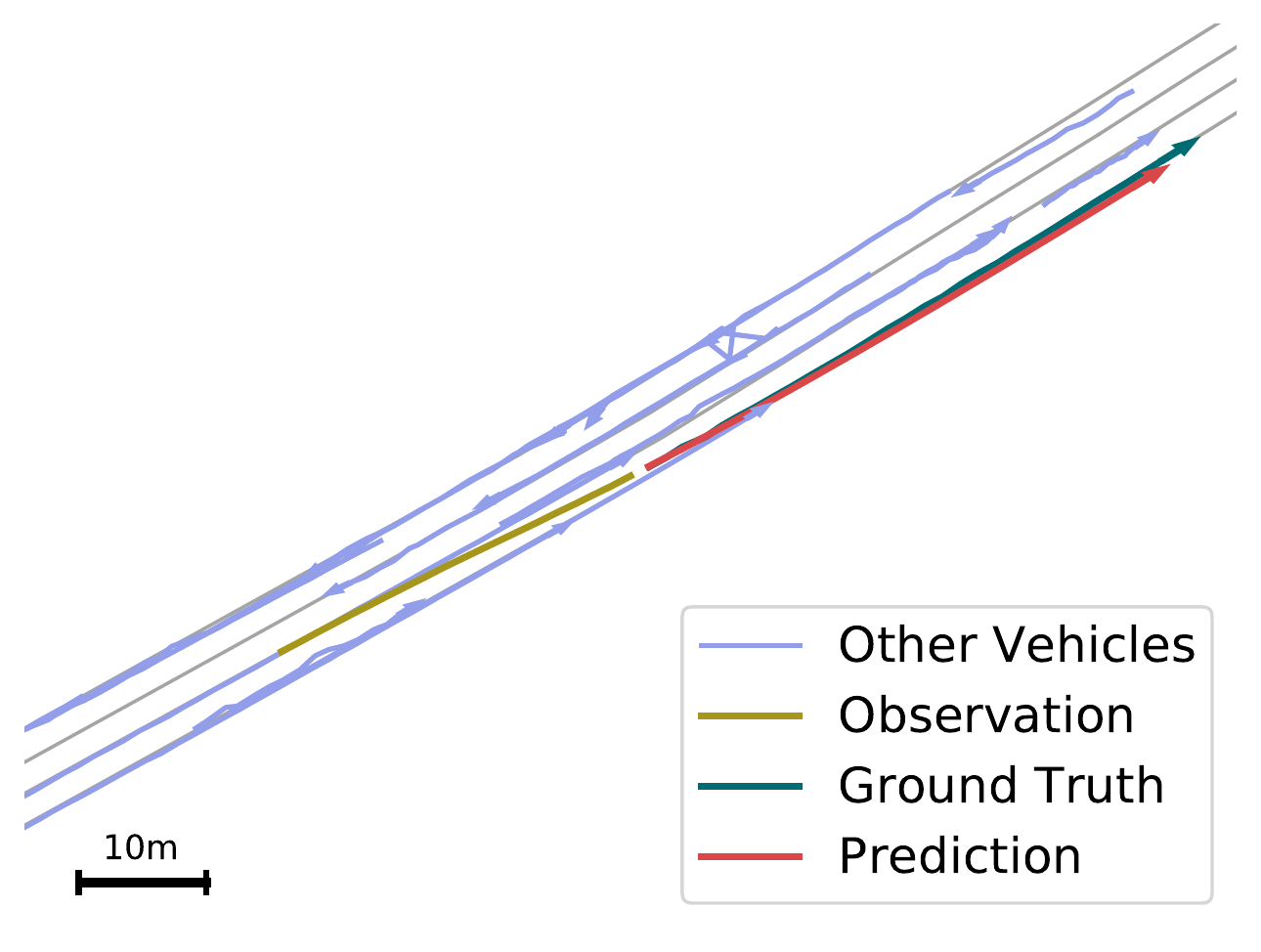}
    \caption{ }
    \end{subfigure}
    \caption{\textbf{Some examples showing the noise in the drivable area map.} All these predictions were considered as off-road because of an inaccurate drivable area map. }
\label{fig:noise-issue}
\end{figure*}

    \item \textbf{Gifs.}
We provide gifs on the perfromance of model when smoothly transforming the scene in \Cref{fig:gifs}. We observe that in some cases the model fails and in some succeeds.

\begin{figure*}[h]
    \centering
\begin{frame}{}
  \animategraphics[loop,controls,width=0.327\linewidth]{10}{supp/smooth/foo-}{0}{28}
  \animategraphics[loop,controls,width=0.327\linewidth]{10}{supp/double/double-}{0}{28}
  \animategraphics[loop,controls,width=0.327\linewidth]{10}{supp/ripple/ripple-}{0}{28}
\end{frame}
    \caption{\textbf{The animations show the changes of the models predictions in different scenes.} It is best viewed using Adobe Acrobat Reader. 
    }
    \label{fig:gifs}
\end{figure*}

\end{enumerate}

\section{Additional quantitative results}
\subsection{Excluding trivial scenes.}
\label{sec:appendix:quant}
In this section, we remove some trivial scenes, i.e., the scenes that fooling is near impossible, e.g., the scenes with zero velocity. Excluding them, we report in \Cref{tab:all_methods-supp} and compared to table 1 of the paper, the off-road numbers substantially increase.

\begin{table*}[!ht]
\begin{center}
\begin{tabular}{|l|c|c|c|c|c|}
\hline
\multirow{3}{4em}{Model} & Original & \multicolumn{4}{c|}{Generated (\textbf{Ours})} 
\\ 
 &&
 \multicolumn{1}{c|}{Smooth-turn} & \multicolumn{1}{c|}{Double-turn} & \multicolumn{1}{c|}{Ripple-road} & 
 \multicolumn{1}{c|}{All} \\
& SOR / HOR   & SOR / HOR  & SOR / HOR   & SOR / HOR   & SOR / HOR  \\
\hline\hline
DATF~\cite{park2020diverse} & 1 / 2 &  44 / 92 & 43 / 91 & 50 / 95 & 51 / 99\\
WIMP~\cite{khandelwal2020whatif} & 0 / 1 &  30 / 80 & 23 / 71 & 29 / 77 & 31 / 82 \\
LaneGCN~\cite{liang2020lanegcn} & 0 / 1 & 23 / 65 & 32 / 75 & 34 / 77 & 37 / 81 \\
MPC~\cite{Ziegler2014benz} & 0 / 0 & 0 / 0 & 0 / 0 & 0 / 0 & 0 / 0 \\

\hline
\end{tabular}
\caption{\textbf{Comparing the performance of different baselines in the original dataset scenes and our generated scenes after removing trivial scenarios.} SOR and HOR are reported in percent and the lower represent a better reasoning on the scenes by the model. Numbers are rounded to the nearest integer.
}
\label{tab:all_methods-supp}
\end{center}
\end{table*}

\subsection{Exploring black box algorithms}
\label{sec:appendix:black}
In the paper, we mentioned that we used a brute-force approach for finding the optimal values as the search space is not huge.
Here, we investigate different block box algorithms for the search. 
The results of applying different search algorithms are provided in \Cref{tab:blackbox}. They cannot overcome the brute-force approach because of their bigger search spaces (the continuous space instead of the discrete space) and the large required computation time.

\begin{table*}[t]
\begin{center}
\begin{tabular}{|l|c|c|}
\hline
\multirow{2}{10em}{Optimization method} & on LaneGCN~\cite{liang2020lanegcn}     & \multirow{2}{5em}{GPU hours} \\ 
& SOR / HOR & \\
\hline\hline

Baysian~\cite{baysian1, baysian2} & 13 / 40 & 17.5 \\
GA~\cite{evolution2} & 14 / 45 & 25.0 \\ 
TPE~\cite{parzen} & 14 / 45 & 12.1 \\ 
Brute force & 23 / 66 & 4.2 \\

\hline
\end{tabular}
\caption{\textbf{Comparing the performance and computation time of different optimization algorithms in the generated scenes.} 
}
\label{tab:blackbox}
\end{center}
\end{table*}

\section{Generalization to rasterized scene}
\label{sec:appendix:raster}
In the paper, we assumed $S$ is in the vector representation, i.e., it includes x-y coordinates of road lanes points. In the case of a rasterized scene, an RGB value is provided for each pixel of the image. Therefore, it is the same as the vector representation unless here we have information (RGB value) about other parts of the scene in addition to the lanes. Hence, the transformation function can be applied directly on all pixels of the image. In other words, in image representation, $s$ is the coordinate of each pixel which has an RGB value and $\hat{s}$ represents the new coordinate with the same RGB value as $s$.

\end{document}